\documentclass{article}

\usepackage{arxiv}

\usepackage[utf8]{inputenc} 
\usepackage[T1]{fontenc}    
\usepackage{hyperref}       
\usepackage{url}            
\usepackage{booktabs}       
\usepackage{amsfonts}       
\usepackage{nicefrac}       
\usepackage{microtype}      
\usepackage{lipsum}

\usepackage{multicol}

\usepackage{url}
\usepackage{xcolor}

\usepackage{hyperref}

\usepackage{makeidx}  
\usepackage{setspace}
\usepackage{amsmath,amsfonts}
\usepackage{microtype}
\usepackage{hyperref}
\usepackage{graphicx}
\usepackage{color}
\usepackage{bbm} 
\usepackage{amssymb}

\newcommand{\algname}{{VoxelMorph}}

\newcommand{\bC}{\boldsymbol{C}}
\newcommand{\bS}{\boldsymbol{S}}
\newcommand{\bD}{\boldsymbol{D}}

\newcommand{\bL}{\boldsymbol{L}}
\newcommand{\bI}{\boldsymbol{I}}
\newcommand{\bw}{\boldsymbol{w}}
\newcommand{\bp}{\boldsymbol{p}}
\newcommand{\br}{\boldsymbol{r}}
\newcommand{\bG}{\boldsymbol{G}}
\newcommand{\bs}{\boldsymbol{s}}

\newcommand{\bA}{\boldsymbol{A}}
\newcommand{\ba}{\boldsymbol{a}}
\newcommand{\bb}{\boldsymbol{b}}

\newcommand{\bmu}{\boldsymbol{\mu}}

\newcommand{\bphi}{\boldsymbol{\phi}}
\newcommand{\bpsi}{\boldsymbol{\psi}}
\newcommand{\bSigma}{\boldsymbol{\Sigma}}

\newcommand{\bLambda}{\boldsymbol{\Lambda}}

\newcommand{\KL}{\text{KL}}
\newcommand{\surfdist}{\text{sd}}

\newcommand{\bz}{\boldsymbol{z}}

\newcommand{\bv}{\boldsymbol{v}}

\newcommand{\bu}{\boldsymbol{u}}

\newcommand{\bmoving}{\boldsymbol{m}}
\newcommand{\bfixed}{\boldsymbol{f}}

\newcommand{\moving}{m}
\newcommand{\fixed}{f}

\newcommand{\Expect}{{\rm I\kern-.3em E}}

\definecolor{newcolor}{rgb}{.8,.349,.1}
\definecolor{blue}{rgb}{0, 0, 0}

\newcommand{\citep}{\cite}

\title{Unsupervised Learning  \\ of Probabilistic Diffeomorphic Registration \\ for Images and Surfaces}%

\author{
		Adrian V. Dalca\\
	MIT and MGH\\
	{\tt\small adalca@mit.edu}\\
	\And
	Guha Balakrishnan\\
	MIT\\
	{\tt\small balakg@mit.edu}\\
	\And
	John Guttag\\
	MIT\\
	{\tt\small guttag@mit.edu}
	\And
	Mert R. Sabuncu\\
	Cornell University\\
	{\tt\small msabuncu@cornell.edu}
}

\begin{document}

\twocolumn[{%
	\begin{@twocolumnfalse}
		\maketitle
		\begin{abstract}


Classical deformable registration techniques achieve impressive results and offer a rigorous theoretical treatment, but are computationally intensive since they solve an optimization problem for each image pair. Recently, learning-based methods have facilitated fast registration by learning spatial deformation functions. 
However, these approaches use restricted deformation models, require supervised labels, or do not guarantee a diffeomorphic (topology-preserving) registration. Furthermore, learning-based registration tools have not been derived from a probabilistic framework that can offer uncertainty estimates.

In this paper, we build a connection between classical and learning-based methods. We present a probabilistic generative model and derive an unsupervised learning-based inference algorithm that uses insights from classical registration methods and makes use of recent developments in convolutional neural networks (CNNs). We demonstrate our method on a 3D brain registration task for both images and anatomical surfaces, and provide extensive empirical analyses. Our principled approach results in state of the art accuracy and very fast runtimes, while providing diffeomorphic guarantees. Our implementation is available at \url{http://voxelmorph.csail.mit.edu}.


		\end{abstract}
	\vspace{0.5cm} 
	\end{@twocolumnfalse}
}]
	
%
%
%


\section{Introduction}
\label{sec:introduction}

Deformable registration computes a dense correspondence between two images, and is fundamental to many medical image analysis tasks. Classical registration techniques have been rigorously developed and studied, but require computationally intensive optimization for each image pair, often requiring tens of minutes to hours of compute time on a CPU. Recent, learning-based registration methods achieve fast runtimes by building on machine learning developments, but largely omit rigorous theoretical treatment of deformations and topology-preserving guarantees. In this work, we present an approach that builds on the strengths of both paradigms, and overcomes these shortcomings. We provide a rigorous connection between probabilistic generative models for deformations and learning algorithms based on convolutional neural networks (CNNs). We also demonstrate that the learning can be done end-to-end in an unsupervised fashion for this model. The resulting framework provides registration for a new image pair in under a second on a GPU, while maintaining guarantees developed for classical methods. 
 
Our formulation casts registration as variational inference on a probabilistic generative model. This framework naturally results in an algorithm that leverages a collection of images to learn a global convolutional neural network with an intuitive cost function. Importantly, we introduce \textit{diffeomorphic integration} layers combined with a spatial transform layer to enable unsupervised end-to-end learning for diffeomorphic registration. We demonstrate that our algorithm achieves state-of-the-art registration accuracy while providing diffeomorphic deformations and fast runtime, and can estimate of registration uncertainty. In our experiments we focus on the example of registering 3D MR brain scans, using a multi-study dataset of over 3,500 scans. However, the method is broadly applicable to many registration tasks.

This paper extends a preliminary version of this work presented at the Medical Image Computing and Computer Assisted Intervention (MICCAI) 2018 conference~\citep{dalca2018pd}. We build on that work by providing theoretical extensions, new results, analysis, and discussion. We first expand the model, including a natural extension to anatomical surfaces. In our experiments, we add baselines, new experiments on registration of both images and surfaces, and provide an analysis of the effect of our diffeomorphic implementation on field regularity and runtime. 
We implement our method as part of the registration framework called VoxelMorph, which is available at \url{http://voxelmorph.csail.mit.edu}.

\subsection{Related Works}

\subsubsection{Classical Registration Methods}
Classical methods solve an optimization over the space of deformations~\citep{ashburner2007,avants2008,bajcsy1989,beg2005,dalca2016,glocker2008,thirion1998, yeo2010learning, zhang2017}. Common representations are displacement vector fields, including elastic-type models~\citep{bajcsy1989, davatzikos1997, shen2002}, free-form deformations with b-splines~\citep{rueckert1999}, statistical parametric mapping~\citep{ashburner2000}, Demons~\citep{pennec1999, thirion1998}, and more recently  discrete methods~\citep{dalca2016, heinrich2013mrf, glocker2008}.

Constraining the allowable transformations to diffeomorphisms ensures certain desirable properties, such as preservation of topology. Diffeomorphic transforms have seen extensive methodological development, yielding state-of-the-art tools, such as Large Diffeomorphic Distance Metric Mapping (LDDMM)~\citep{beg2005,cao2005large,ceritoglu2009multi, hernandez2009registration, joshi2000landmark, miller2005increasing, oishi2009atlas, zhang2017}, DARTEL~\citep{ashburner2007}, diffeomorphic Demons~\citep{vercauteren2009}, and symmetric normalization (SyN)~\citep{avants2008}. In general, these tools demand substantial time and computational resources for a given image pair. 
	
Some recent GPU-based iterative algorithms use these frameworks to develop faster algorithms by requiring a GPU to be available for each registration~\citep{modat2010fast,modat2014global}. Recent learning-based registration methods have demonstrated that they can provide good initializations to iterative GPU methods~\citep{balakrishnan2019} to further improve runtime. 

{\color{blue}Probabilitic image registration methods specify priors on the deformation between two images, and likelihood models that describe image intensities~\citep{simpson2012probabilistic,zhang2017,heinrich2016deformable,risholm2013,amirkhalili2017}. These formulations also lead to iterative optimization methods, but can yield distributions of deformation fields. In this paper, we build on these models by presenting a general variational inference strategy to optimize a global neural network that efficiently outputs distributions of deformations.}

\subsubsection{Learning-based Registration}
Recent methods have proposed to train neural networks that map a pair of input images to an output deformation. Most earlier approaches demonstrated the feasibility of deep learning based registration, and required ground truth registration fields~\citep{cao2017deformable,krebs2017,rohe2017,sokooti2017,yang2017}. Such ground truth deformations are often derived via more conventional registration tools or simulations, sometimes limiting their applicability. 

Building on the successful demonstration of these methods, several recent papers~\citep{balakrishnan2018a,balakrishnan2019,devos2017,devos2019,li2017} explore unsupervised, or end-to-end, strategies. These methods employ a neural network that computes spatial transformation~\citep{jaderberg2015} to warp one image to another, enabling end-to-end training. A recent approach builds on these methods by learning a spatially-adaptive regularizer within a
registration model~\citep{niethammer2019} These approaches use machine learning techniques to achieve efficient training and fast runtimes, but build on classical registration development, such as probabilistic models and diffeomorphic theory. In our work, we bridge these two paradigms to offer classical guarantees within a machine learning approach.
{\color{blue}We note the contemporaneous development of a method that uses a conditional variational auto encoder (CVAE) to learn diffeomorphic representations~\citep{krebs2018unsupervised,krebs2019learning}. Similar to our method, this approach uses a variational strategy to learn a network to predict a stationary velocity field (SVF). However, the authors focus on representing the SVF through the manifold of the CVAE, and focus on the anatomical variation captured through this encoding}

Recent methods proposed using segmentation-based cost functions, such as Dice~\citep{dice1945}, to replace the image-based similarity term when segmentations are available during training for multi-modal registration, such as T2w MRI and 3D ultrasound, within the same subject~\citep{hu2018weakly,hu2018label}. We extend this line of work by showing that our generative probabilistic model naturally describes the deformation of surfaces, therefore enabling the use of segmentations during training within a single cohesive framework. The extended model results in a combination of segmentation (surface) and image-based training losses.



{\color{blue}
\subsubsection{\color{blue}Surface-based Registration}
	In this paper, we also present an extension to our main contribution which enables alignment of surfaces. In medical image registration, surface matching methods often use surface coordinates or geometric features extracted from anatomical structures~\citep{aiger2008,durrleman2010,miga2003,postelnicu2008}. Several methods treat surfaces as 3D point sets with shape descriptors, and often use Iterated Closest Point based optimization methods to find the shape correspondences~\citep{besl1992method}. Currents, defined as unconnected oriented points, have been used to register surfaces, for example using Matching Pursuit algorithms~\cite{durrleman2010}. Some methods combine volume and surface registration, often using surface registrations to initialize dense volumetric transforms~\cite{postelnicu2008}. Similar to volume registration, these classical surface matching methods use iterative optimization strategies, requiring significant computational resources. Building on these methods, we use a 3D point representation jointly with images to achieve fast registration with neural networks, enabled by a differentiable surface distance function.}

\subsection{Background: Diffeomorphic Registration}

Although the method presented in this paper applies to a multitude of deformable representations, we choose to work with diffeomorphisms, and in particular with a stationary velocity field representation~\citep{ashburner2007}.  Diffeomorphic deformations are differentiable and invertible, and thus preserve topology. Let~\mbox{$\bphi: \mathbb{R}^3 \rightarrow \mathbb{R}^3$} represent the deformation that maps the coordinates from one image to coordinates in another image. 
In our implementation,  the deformation field is defined through the following ordinary differential equation (ODE): 
\begin{equation}
\frac{\partial\bphi^{(t)}}{\partial t} = \bv(\bphi^{(t)})
\label{eq:ode}
\end{equation}
%
%
where $\bphi^{(0)} = Id$ is the identity transformation and~$t$ is time. We integrate  the stationary velocity field $\bv$ over $t=[0,1]$ to obtain the final registration field $\bphi^{(1)}$~\citep{moler2003}.

While we implement and evaluate several numerical integration techniques, we find \textit{scaling and squaring} to be most efficient, and we briefly review the technique here~\citep{arsigny2006}.
The integration of a stationary ODE represents a one-parameter subgroup of diffeomorphisms.  In group theory, $\bv$ is a member of the Lie algebra and is exponentiated to produce $\bphi^{(1)}$, which is a member of the Lie group: $\bphi^{(1)} =\text{exp}(\bv)$.
  From the properties of one-parameter subgroups, for any scalars $t$ and $t'$, \mbox{$\exp((t+t')\bv) = \exp(t\bv) \circ \exp(t'\bv)$}, where $\circ$ is a composition map associated with the Lie group. 
Starting from~$\bphi^{(1/2^T)} =\bp + \bv(\bp)/2^T$ where~$\bp$ is a map of spatial locations, we use the recurrence~$\bphi^{(1/2^{t-1})} = \bphi^{(1/2^t)} \circ \bphi^{(1/2^t)}$ to obtain~$\bphi^{(1)} = \bphi^{(1/2)} \circ \bphi^{(1/2)}$. $T$ is chosen so that $\bv/2^T$ is very small.



\begin{figure}[t!]
	\centering
		\includegraphics[width=1\linewidth]{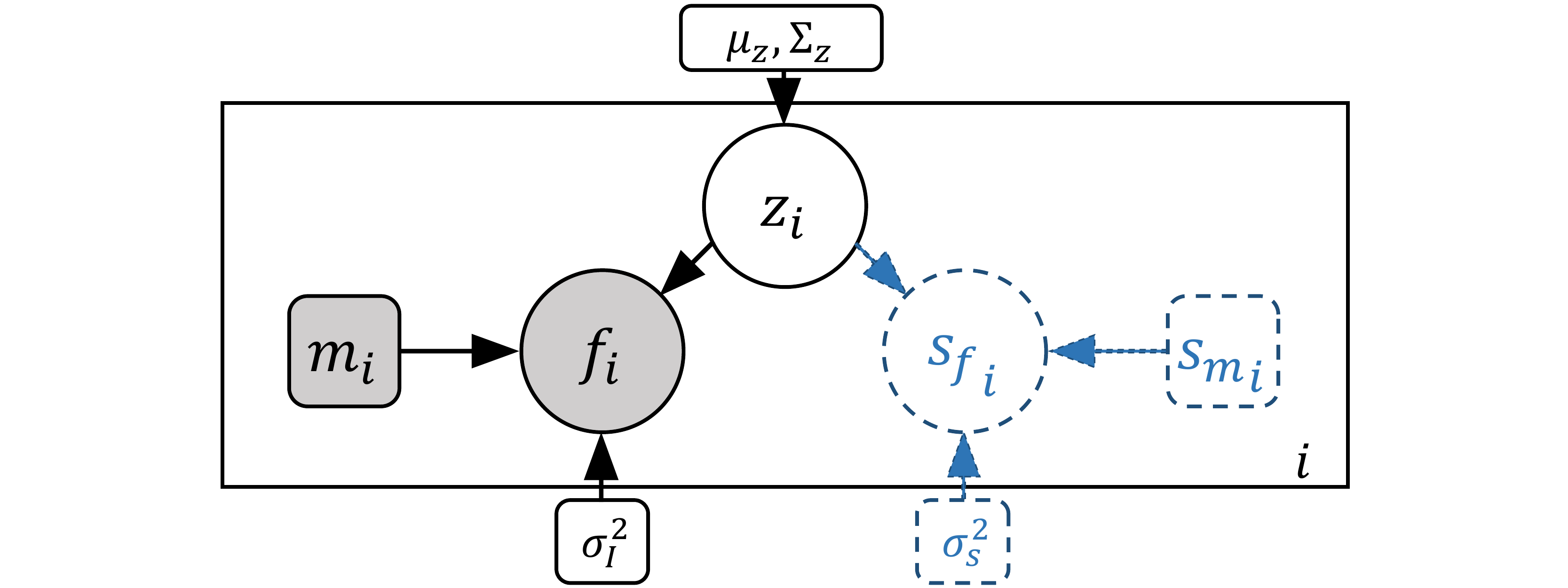}
	\caption{A graphical representation of our generative model. Circles indicate random variables and rounded squares represent parameters. Shaded quantities are observed \textit{at test time}, and the plates indicate replication.~$\bfixed$ and~$\bmoving$ are the input images. The image intensities~$\bfixed$ are generated from a normal distribution centered at~$\bmoving \circ \bphi_z$. The registration prior is defined by normal parameters~$\bmu_{z}$, and~$\bSigma_{z}$. In blue, the \textit{optional} similar model structure is included for an anatomical surface, used purely for learning an improved posterior of registration.}
	\label{fig:graphicalmodel_simple}
\end{figure}

\begin{figure*}[t!]
	\centering
	\begin{center}
	  \includegraphics[width=1.0\linewidth]{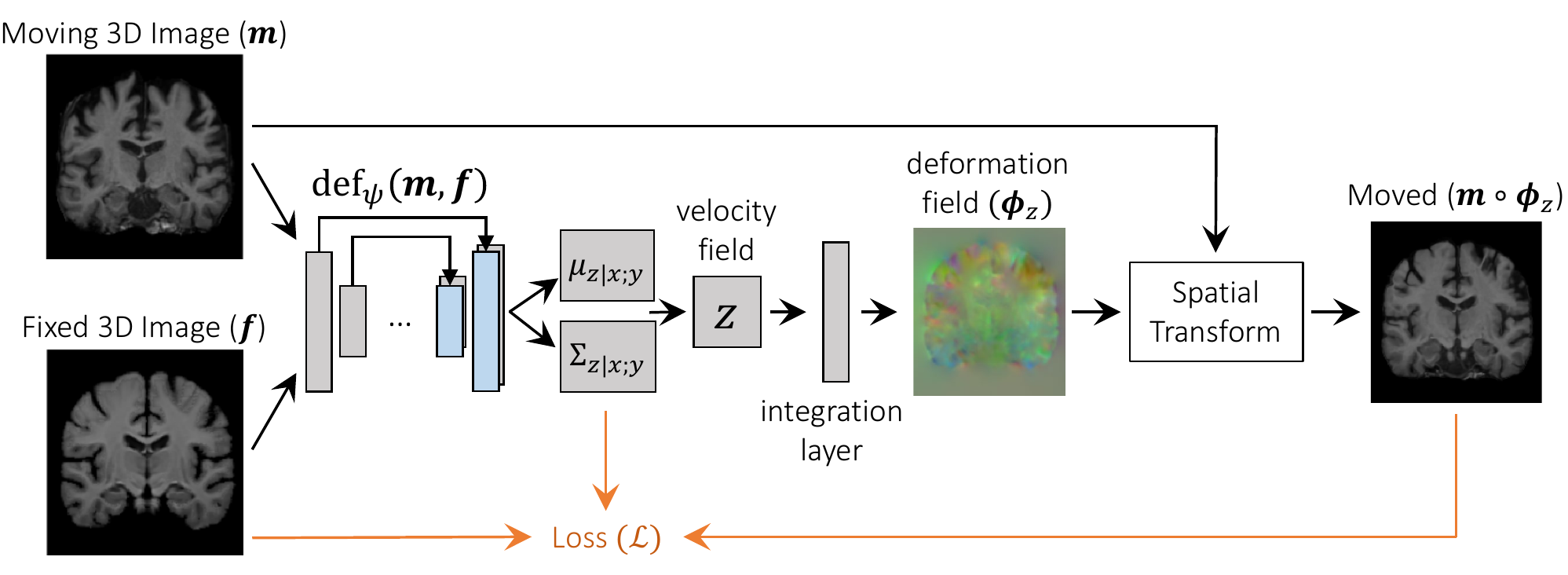}
	\end{center}
	\hfill
	\begin{minipage}[b]{1\linewidth}
		\caption{Overview of end-to-end unsupervised architecture. The first part of the network,~$\text{def}_{\psi}(\bmoving, \bfixed)$ takes the input images and outputs the approximate posterior probability parameters representing the velocity field mean,~$\bmu_{z|\moving;\fixed}$, and variance,~$\bSigma_{z|\moving;\fixed}$. A velocity field~$\bz$ is sampled and transformed to a diffeomorphic deformation field~$\bphi_z$ using novel differentiable \textit{squaring and scaling} integration layers. Finally, a spatial transform warps~$\bmoving$ to obtain~$\bmoving \circ \bphi_z$. Figure~\ref{fig:sup:network_overview_simple} expands on this overview by including the optional surface-based loss.
		}
		\label{fig:network_overview_simple}
	\end{minipage}
\end{figure*}

\vspace{-0.1cm}
\section{Methods}
\label{sec:model}

We let~$\bfixed$ and~$\bmoving$ be 3D images, such as MRI volumes,   
%
%
and let~$\bz$ be a latent variable that parametrizes a transformation function~$\bphi_{\bz}: \mathbb{R}^3 \rightarrow \mathbb{R}^3$. We propose a generative model that describes the formation of~$\bfixed$ by warping~$\bmoving$ via~$\bmoving \circ \bphi_{\bz}$. 
%
%
We propose a variational inference approach that leverages a convolutional neural network with diffeomorphic integration and spatial transform layers. We learn network parameters in an unsupervised fashion, without access to ground truth registrations. We describe how the network yields fast diffeomorphic registration of a new image pair~$(\bfixed,\bmoving)$, in a probabilistic framework. We expand this treatment by including anatomical surface alignment, which enables training the network given (optional) anatomical segmentations.

\subsection{Generative Model}

We model the prior probability of the parametrization~$\bz$ as:
\begin{align}
p(\bz) = \mathcal{N}(\bz; \boldsymbol{0}, \bSigma_z),  
\end{align}
where~$\mathcal{N}(\cdot;\bmu,\bSigma)$ is the multivariate normal distribution with mean~$\bmu$ and covariance~$\bSigma$. 
Our work applies to a wide range of representations~$\bz$. 
For example,~$\bz$ could be a dense displacement field, or a low-dimensional embedding of the displacement field. In this paper, we let~$\bz$ be a stationary velocity field that specifies a diffeomorphism through the ODE~\eqref{eq:ode}. We let~$\bL = \bD - \bA$ be the Laplacian of a neighborhood graph defined on the voxel grid, where~$\bD$ is the graph degree matrix, and~$\bA$ is a voxel neighbourhood adjacency matrix. We encourage \textit{spatial smoothness} of the velocity field~$\bz$ by setting~$\bSigma_z^{-1} = \bLambda_z = \lambda \bL$, where~$\bLambda_z$ is a precision matrix and~$\lambda$ denotes a parameter controlling the scale of the velocity field~$\bz$.

We let~$\bfixed$ be a noisy observation of warped image~$\bmoving$:
\begin{align}
	p(\bfixed | \bz ; \bmoving) &= \mathcal{N}(\bfixed ; \bmoving \circ \bphi_{\bz}, \sigma_I^2\boldsymbol{\mathbbm{I}}),  
	\label{eq:likelihood}
\end{align}
where~$\sigma_I^2$ captures the variance of additive image noise.

We aim to estimate the posterior registration probability~$p(\bz|\bfixed;\bmoving)$. Using this, we can obtain the most likely registration field~$\bphi_{\bz}$ for a new image pair~$(\bfixed, \bmoving)$ via MAP estimation, along with an estimate of velocity field variance at each voxel. Figure~\ref{fig:graphicalmodel_simple} provides a graphical representation of our model.

\vspace{-0.1cm}
\subsection{Learning}

Given our assumptions, computing the posterior probability~$p(\bz|\bfixed; \bmoving)$ is intractable. We use a variational approach, and introduce an approximate posterior probability~$q_{\bpsi}(\bz|\bfixed; \bmoving)$ parametrized by~$\bpsi$. We minimize the KL divergence
%
\begin{align}
&\min_{\psi} \KL \left[q_{\bpsi}(\bz|\bfixed ; \bmoving) || p(\bz|\bfixed ; \bmoving)  \right] \nonumber \\
&= \min_{\psi} \Expect_{q} \left[ \log q_{\bpsi}(\bz|\bfixed ; \bmoving) - \log p(\bz|\bfixed ; \bmoving) \right] \nonumber \\
&= \min_{\psi} \Expect_{q} \left[ \log q_{\bpsi}(\bz|\bfixed ; \bmoving) - \log p(\bz, \bfixed; \bmoving) \right] + \log p(\bfixed ; \bmoving) \nonumber \\
&= \min_{\psi} \KL \left[  q_{\bpsi}(\bz|\bfixed; \bmoving) ||  p(\bz)  \right] - \Expect_{q} \left[ \log p(\bfixed | \bz ; \bmoving) \right] + \text{const},
\label{eq:VLB}
\end{align}
%
%
%
%
which yields the negative of the \textit{variational lower bound} of the model evidence~\citep{kingma2013}.
%
We model the approximate posterior~$q_{\bpsi}(\bz | \bfixed ; \bmoving)$ as a multivariate normal:
\begin{align}
q_{\bpsi}(\bz | \bfixed ; \bmoving) = \mathcal{N}(\bz ; \bmu_{z | \moving, \fixed}, \bSigma_{z |\moving, \fixed}),
\end{align}
where we let~$\bSigma_{z | \moving, \fixed}$ be diagonal. To understand the effects of this assumption, we explore a non-diagonal covariance in a later section. The statistics~$\bmu_{z | \moving, \fixed}$ and the diagonal of $\bSigma_{z |\moving, \fixed}$ can be interpreted as the voxel-wise mean and variance, respectively.

We estimate~$\bmu_{z | \moving, \fixed}$, and~$\bSigma_{z | \moving, \fixed}$ using a convolutional neural network~$\text{def}_{\bpsi}(\bfixed,\bmoving)$ parameterized by~$\bpsi$, as described in the next section.
%
We learn parameters~$\bpsi$ by optimizing the variational lower bound~\eqref{eq:VLB} using stochastic gradient methods. Specifically, for each image pair~$(\bfixed, \bmoving)$ and sample~$\bz_k\sim q_{\psi}(\bz|\bfixed; \bmoving)$, we compute~$\bmoving \circ \bphi_{z_k}$,
with the resulting loss (detailed derivation in supplementary material):
\begin{align}
&\mathcal{L}(\bpsi; \bfixed, \bmoving)  = - \Expect_{q} \left[ \log p(\bfixed | \bz ; \bmoving) \right] \nonumber\\
&+ \KL \left[  q_{\bpsi}(\bz|\bfixed; \bmoving) ||  p(\bz)  \right]   \nonumber \\
&= \frac{1}{2\sigma^2K} \sum_k ||\bfixed - \bmoving \circ \bphi_{z_k}||^2 \nonumber \\
&+ \frac{1}{2} \left[ \text{tr}(\lambda\bD \bSigma_{z|x;y} - \log\bSigma_{z|x;y}) + \bmu_{z | \moving, \fixed}^T \bLambda_z \bmu_{z | \moving, \fixed} \right] \nonumber \\
&+ \text{const}, 
\label{eq:main_loss}
\end{align}
where~$K$ is the number of samples used to approximate the expectation. 
The first term encourages image~$\bfixed$ to be similar to the warped image~$\bmoving \circ \bphi_{z_k}$. The second term encourages the posterior to be close to the prior~$p(\bz)$. Although the variational covariance~$\bSigma_{z|\moving,\fixed}$ is diagonal, the last term spatially smoothes the mean, which can be seen by expanding
%
~$\bmu_{z | \moving, \fixed}^T \bLambda_z \bmu_{z | \moving, \fixed} = \frac{\lambda}{2} \sum \sum_{j\in N(I)} (\bmu[i] - \bmu[j])^2$,
%
where
%
$N(i)$ are the neighbors of voxel~$i$. 
We treat~$\sigma^2$ and~$\lambda$ as fixed hyper-parameters that we investigate in our experiments, and use~$K=1$.

\subsection{Neural Network Framework}

We design the network~~$\text{def}_{\bpsi}(\bfixed,\bmoving)$ that takes as input~$\bfixed$ and~$\bmoving$ and outputs~$\bmu_{z|\moving,\fixed}$ and~$\bSigma_{z|\moving,\fixed}$, based on a 3D UNet-style architecture~\citep{ronneberger2015}. 
{\color{blue}The network includes a convolutional layer with 32 filters, four downsampling layers with 64 convolutional filters and a stride of two, and three upsampling convolutional layers with 64 filters. We only upsample three times to predict the velocity field (and following integration steps) at every two voxels, to enable these operations to fit in current GPU card memory.}

To enable unsupervised learning of parameters~$\bpsi$ using~\eqref{eq:main_loss}, we must form~\mbox{$\bmoving \circ \bphi_z$} and compute the data term. We first implement a layer that samples a new \mbox{$\bz_k \sim \mathcal{N}(\bmu_{z|\moving,\fixed}, \bSigma_{z|\moving,\fixed})$} using the ``re-parameterization trick"~\citep{kingma2013}: $\bz_k = \bmu_{z | \moving, \fixed} + \sqrt{\bSigma_{z | \moving, \fixed}} \br$, where~$\br$ is a sample from the standard normal:~$\br \sim \mathcal{N}(0, \bI)$.


Given~$\bz_k$, we need to compute~\mbox{$\bphi_{z_k} = \exp(\bz_k)$} as described in the introduction. We propose vector integration layers using \emph{scaling and squaring} operations.  
%
%
Specifically, \emph{scaling and squaring} operations involve compositions within the neural network architecture using a differentiable spatial transformation operation. Given two 3D vector fields $\ba$ and $\bb$, for each voxel $p$ this operation computes $(\ba \circ \bb)(p) = \ba(\bb(p))$, a non-integer voxel location $\bb(\bp)$ in $\ba$, using linear interpolation. Starting with~$\bphi^{(1/2^T)} = \bp + \bz_k/2^T$, we compute~\mbox{$\bphi^{(1/2^{t-1})} = \bphi^{(1/2^t)} \circ \bphi^{(1/2^t)}$} recursively using these operations T times, leading to~$\bphi^{(1)} \triangleq \bphi_{z_k} = \exp(\bz_k)$. In our experiments, we extensively analyze the effect of the step size~$T$ on the runtime of the network, the accuracy of the registration, and the regularity of the deformation. We also implement vector integration layers using quadrature and ODE solvers, and in the experiments show that these are significantly slower and can require significant memory.

Finally, we warp volume~$\bmoving$ according to the computed diffeomorphic field~$\bphi_{z_k}$ using a spatial transform layer. 

In summary, the network takes as input images~$\bfixed$ and~$\bmoving$, computes statistics~$\bmu_{z|\moving,\fixed}$ and~$\bSigma_{z|\moving,\fixed}$, samples a new velocity field~$\bz_k \sim \mathcal{N}(\bmu_k, \bSigma_k)$, computes a diffeomorphic~$\bphi_{z_k}$ and warps~$\bmoving$. Since all the steps are designed to be differentiable, we  learn the network parameters using stochastic gradient descent-based methods. This network results in three outputs,~$\bmu_{z|\moving,\fixed}, \bSigma_{z|\moving,\fixed}$ and~$\bmoving \circ \bphi_{z_k}$, which are used in the model loss~\eqref{eq:main_loss}.
 The framework is summarized in Figure~\ref{fig:network_overview_simple}.


\subsection{Registration}

Given learned parameters, we approximate registration of a new scan pair~$(\bfixed, \bmoving)$ using~$\bphi_{\hat{z}_k}$. 
We first obtain the most likely velocity field~$\hat{\bz}_k$ using
\begin{align}
\hat{\bz}_k &= \arg \max_{\bz_k} p(\bz_k | \bfixed; \bmoving) = \bmu_{z|\moving;\fixed},
\label{eq:MAP1}
\end{align}
by evaluating the neural network~$\text{def}_\psi(\bfixed, \bmoving)$. We then compute~$\bphi_{\hat{z}_k}$ using the \textit{scaling and squaring} based integration, altogether requiring less than a second on a GPU. We highlight that at test time, the diagonal covariance~$\bSigma_{z | \moving, \fixed}$ is not used, however it enables an estimation of the deformation uncertainty. Analysis of uncertainty is beyond the scope of this paper, and is an interesting avenue for future study.

Using a stationary velocity field representation, computing the inverse deformation field~$\bphi^{-1}_{\bz}$ can be achieved by integrating the negative of the velocity field:~$\bphi^{-1}_{\bz} = \bphi_{-\bz}$, since~$\bphi_{\bz} \circ \bphi_{-\bz} = \exp(\bz)\circ\exp(-\bz) = \exp(\bz-\bz) = Id$~\citep{ashburner2007,modat2014}. This enables the computation of both fields inside one efficient network when desired.

{\color{blue}
\subsection{Implementation}

We implement our method as part of the VoxelMorph package~\citep{balakrishnan2018a}, available online at~\mbox{\url{http://voxelmorph.csail.mit.edu}}, using neuron~\citep{dalca2018priors} and Keras~\citep{chollet2015} with a Tensorflow~\citep{abadi2016} backend. We use a learning rate of~$1e-4$ for the Adam optimizer~\cite{kingma2014}, a batch size of 1 due to memory constraints, and Glorot uniform initialization for the convolution weights. We use a single sample ($K=1$), which has been shown to lead to useful gradients for optimization while maintaining the memory footprint and implementation complexity low~\citep{kingma2013}. For large volumes, the number of samples is often constrained by the available GPU memory. }

\section{Method Extensions}

\subsection{Surface-based Registration}

In various instances, anatomical segmentation maps for specific structures of interest may also be available with some of the training images. Recent papers have demonstrated that the use of segmentations can help in registration~\citep{balakrishnan2019,hu2018label}. Here, we show that our proposed model naturally extends to handle surfaces, enabling the use of segmentations during training within the same principled framework. 

We focus on the case where one anatomical structure is segmented in the image. Given a segmentation map where each voxel is assigned the desired anatomical label or background, we extract the anatomical surface and let $\bs_{\bfixed}$ represent the~$N$ surface \textit{coordinates} of the anatomical structure for image~$\bfixed$, which can be stored as an~$N \times 3$ matrix. Given the diffeomorphism~$\bphi_z$ in the previous section, we model each surface location~$\bs_f[n]$, as formed by displacing a matching surface location~$\bs_m[n]$ according to~$\bphi_z$, and adding (spatial) displacement noise:
\begin{align}
p(\bs_{\bfixed} | \bz ; \bs_{\bmoving}) &= \mathcal{N}(\bs_{\bfixed} ; \bs_{\bmoving} \circ \bphi_{\bz}, \sigma_s^2\boldsymbol{\mathbbm{I}}),  
\label{eq:likelihood-surfaces}
\end{align}
where the composition~$\bs_{\bmoving} \circ \bphi_{\bz}$ warps surface coordinates.

\begin{figure}[t!]
	\centering
	\includegraphics[width=1\linewidth]{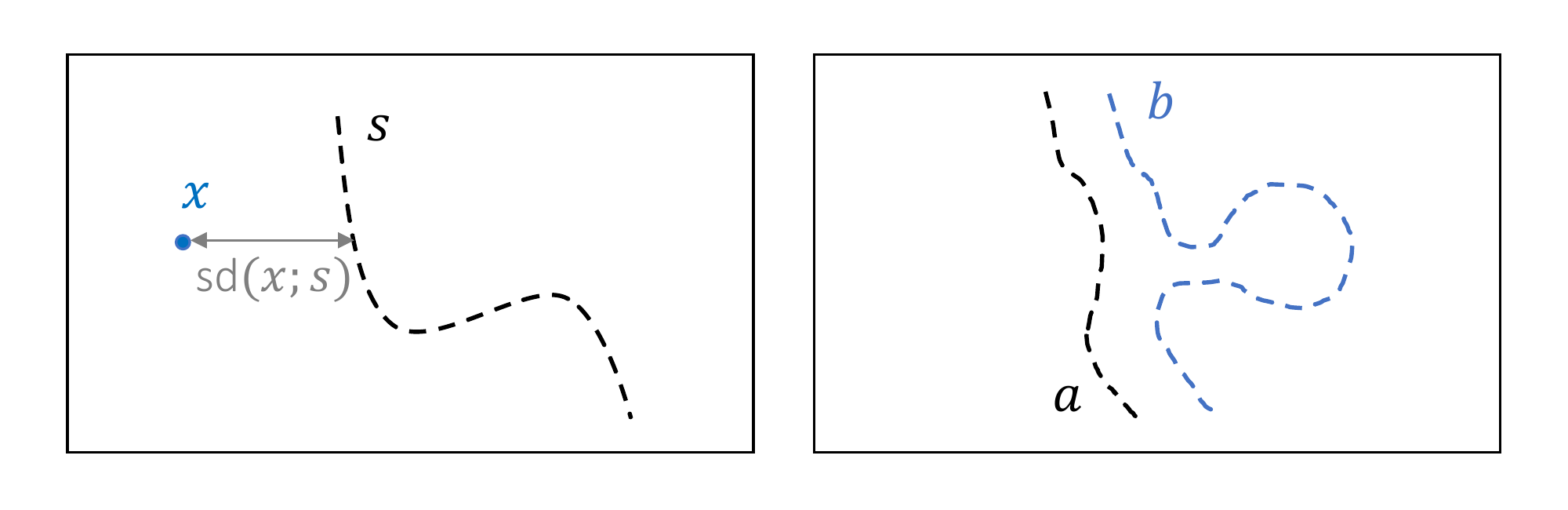}
	\hfill
	\vspace{-0.75cm}
	\caption{\color{blue} Left: an illustration of the surface distance function~$\surfdist(x;\bs)$. Right: asymmetric surface behavior requires that we compute the surface distance in both directions. For example, computing~$\sum_v \surfdist(\ba[n], \bb)$ will be considerably smaller than~$\sum_v sd(\bb[n], \ba)$ due to surface points on the hairpin of $\bb$ (recall that surface points are not directly corresponding.)}
	\label{fig:surface_explanation}
\end{figure}

Given both images and segmentation maps during training, we extract surfaces of the desired structure and aim to estimate the posterior probability~$p(\bz|\bfixed, \bs_{\bfixed}; \bmoving, \bs_{\bmoving})$. As before, we use a variational approximation. Since segmentation maps, and hence surfaces, are usually derived from images, we assume that images are sufficient to approximate the posterior:~\mbox{$q(\bz|\bfixed, \bs_{\bfixed}; \bmoving, \bs_{\bmoving}) = q_{\bpsi}(\bz|\bfixed; \bmoving)$}. As before, we minimize the KL divergence between the true and approximate posterior (derived in supplementary material):
\begin{align}
&\min_{\psi} \KL \left[q_{\bpsi}(\bz|\bfixed ; \bmoving) || p(\bz|\bfixed, \bs_{\bfixed}; \bmoving, \bs_{\bmoving}) \right] \nonumber \\
=&\min_{\psi} \KL \left[  q_{\bpsi}(\bz|\bfixed; \bmoving) ||  p(\bz)  \right] 
- \Expect_{q} \left[ \log p(\bfixed | \bz ; \bmoving) \right] \nonumber \\
&-  \Expect_{q} \left[ \log p(\bs_{\bfixed} | \bz ; \bs_{\bmoving}) \right],
\end{align}
and arrive at the loss function:
\begin{align}
&\mathcal{L}(\bpsi; \bfixed, \bs_{\bfixed}, \bmoving, \bs_{\bmoving})  \nonumber \\
&= \frac{1}{2} \left[ \text{tr}(\lambda\bD \bSigma_{z|x;y} - \log\bSigma_{z|x;y}) + \bmu_{z | \moving, \fixed}^T \bLambda_z \bmu_{z | \moving, \fixed} \right] \nonumber \\
&+ \frac{1}{2\sigma_I^2K} \sum_k ||\bfixed - \bmoving \circ \bphi_{z_k}||^2  \nonumber \\
&+ \frac{1}{2\sigma_s^2K} \sum_k ||\bs_{\bfixed} - \bs_{\bmoving} \circ \bphi_{z_k}||^2.
\label{eq:surf-loss}
\end{align}
Compared to the original model loss~\eqref{eq:main_loss}, the additional third term encourages the deformation~$\bphi_{z_k}$ to warp the moving surface close to the fixed surface~$\bs_f$. As described in the generative model~\eqref{eq:likelihood-surfaces}, this requires \textit{corresponding} surface points in~$\bs_{\bfixed}$ and $\bs_{\bmoving}$. However, these correspondences are not available in practice, as segmentations are provided independently for each image. Therefore, the third term cannot be computed directly.

We propose an approximation of the surface term using surface \textit{distance transforms}. Let~$\surfdist(x, \bs)$ be a \textit{surface distance} function, which for location~$x$ returns the Euclidean distance to the closest surface point in~$\bs$ (Figure~\ref{fig:surface_explanation}Left).\footnote{Function~$\surfdist(x, \bs)$ is a generating function for a distance transform image for the surface~$\bs$, by evaluating it at every grid point~$x$} Noting that for two surfaces~$\ba$ and~$\bb$, $ \sum_n \surfdist(\ba[n], \bb) \neq \sum_n \surfdist(\bb[n], \ba)$ due to potential asymmetries in the surfaces (see Figure~\ref{fig:surface_explanation}Right), we approximate the distance~$\| \bs_{\bfixed} - \bs_{\bmoving} \circ \bphi_{z_k}\|^2$ by computing~$\surfdist(\cdot, \cdot)$ in both directions: 
\begin{align}
&\| \bs_{\bfixed} - \bs_{\bmoving} \circ \bphi_{z_k}\|^2 \nonumber \\
&\approx \frac{1}{2} \sum_n \surfdist(\bs_{\bfixed}[n] \circ \bphi_z^{-1}, \bs_{\bmoving} ) + \sum_n  \surfdist(\bs_{\bmoving} [n] \circ \bphi_z, \bs_{\bfixed} ).
\label{eq:surface-approx}
\end{align}
{\color{blue} We implement this function efficiently using distance transforms. Specifically, to compute~\mbox{$\surfdist(\bs_{\bmoving} [n] \circ \bphi_z, \bs_{\bfixed})$}, we first pre-compute distance transforms for the (fixed) given structure~$\bs_f$. We then sample~$100,000$ points along~$\bs_m$, which we find to be sufficient to estimate accurate measures along the surface. We warp (move) them according to the deformation~$\bphi_z$, and compute the distance transform of~$\bs_f$ at these locations. We take advantage of our diffeomorphic representation that enables computing the inverse~$\bphi_z^{-1}$ efficiently within the network to similarly compute~\mbox{$\surfdist(\bs_{\bfixed} [n] \circ \bphi_z^{-1}, \bs_{\bmoving})$}.}

\begin{table*}[h!]
	\small
	\centering
	\color{blue}
	\begin{tabular}{c c c c c c}
		Method&Avg. Dice&GPU sec&CPU sec & mean $|J_\Phi|$ & $|J_\Phi| \le 0$    \\
		\hline
		Affine only&0.584 (0.157)&0 &0  & 1 & 0  \\
		ANTs (SyN) &0.749 (0.136)&-&9059 (2023) & 1.001 (0.036) & 7523 (4790)\\
		NiftyReg (CC) &0.755 (0.143)&-&2347 (202) & 1.072 (0.131)  & 33838 (8307) \\
		VoxelMorph (CC) & 0.753 (0.145) & 0.45 (0.01) & 57 (1.0) & 1.032 (0.074) & 19715 (3540)   \\ 
		Supervised-diff & 0.730 (0.144) & 0.35 (0.03) & 82.6 (3.8)        & 1.088 (0.121) & 0.05 (0.5)   \\ 
		\hline
		\verb|VoxelMorph-diff| & 0.754 (0.139) & 0.47 (0.01)& 84.2 (0.1) & 1.075 (0.124) & 0.2 (1.0)   \\
		\hline  
		\vspace{-0.5cm}
	\end{tabular}
	\caption{\color{blue} Summary of results: mean Dice scores over all anatomical structures and subjects (higher is better), mean runtime; mean Jacobian determinant; and mean number of locations with a non-positive Jacobian determinants of each registration field (lower is better). All methods have comparable Dice scores, while our method and the other VoxelMorph variants are orders of magnitude faster than ANTs or NiftiReg. Only our presented method, VoxelMorph-diff,
		achieves both high accuracy and fast runtime while also having nearly zero non-negative Jacobian locations. All methods have mean Jacobian determinants close to 1, indicating smooth deformations.
		Each aspect of these results is studied in more details in the rest of the experiments and figures.
	}
	\label{tbl:results}
\end{table*}

In summary, since to compute the posterior approximation~$q_{\bpsi}(\bz|\bfixed; \bmoving)$ the neural network takes as input only the images~$\bfixed$ and~$\bmoving$, images alone are required at test time. Given a diffeomorphism~$\bphi_{z_k}$, at training time the network uses both a warped image and a warped surface to evaluate the quality of the registration. 

This model can also be used to register two surfaces when the images themselves are not available. The only modelling change required is removing the image likelihood terms and using the variational approximation $q_{\bpsi}(\bz|\bS_{\bfixed}; \bS_{\bmoving})$, which uses the segmentation maps~$\bS_m$ and~$\bS_f$ as input. Surface-only registration is beyond the scope of this paper, and we leave it for future work. {\color{blue}However, registration with images and surfaces is described here as an example of possible extensions of the model, and surface-\textit{only} registration is beyond the scope of this paper. }

The complete neural network framework, including the surface loss, is illustrated in supplemental~Figure~\ref{fig:sup:network_overview_simple}.

\subsection{Non-diagonal Covariance}
\label{sec:nondiag-approx}

{\color{blue} Approximating the velocity field covariance~$\bSigma_{z | \moving, \fixed}$ using a diagonal matrix is a strong assumption that ignores spatial smoothness. As seen in~\eqref{eq:main_loss}, the spatially-smooth prior~$p(\bz)$ encourages a smooth mean velocity field~$\bmu_{z | \moving, \fixed}$, but samples~\mbox{$\bz_k \sim \mathcal{N}(\bmu_{z | \moving, \fixed}, \bSigma_{z | \moving, \fixed})$} might still be noisy. In this section, we investigate the effects of this restriction, by providing a model expansion that computes a less restrictive covariance. In our experiments below, we analyze the effects of these different approximations.}

To evaluate the effects of the diagonal covariance, we explore a second approximation~$\bSigma_{z | \moving, \fixed} = \bC_{\sigma_c} \bG \bG^{T} \bC_{\sigma_c}^{T}$ where~$\bG$ is a diagonal matrix returned by the neural network and~$\bC_{\sigma_c}$ is a fixed smoothing convolution matrix.  Specifically, for each row of~$\bC_{\sigma_c}$ we create a flattened Gaussian smoothing kernel centered at a particular voxel, such that~$\bC_{\sigma_c}\bw$ is equivalent to 3D convolution of image~$\bw$ by a gaussian filter with variance~$\sigma_c^2$. We choose~$\sigma_c$ such that the smoothing operation matches the scale of the prior~$p(\bz)$ determined by~$\lambda$:~$\frac{1}{\sqrt{2\pi\sigma_c^{3/2}}} = (\lambda*6)^{-1}$. 

During training, sampling from the posterior is achieved using the \textit{reparametrization trick}:~$\bz_k = \bmu_{z | \moving, \fixed} + \bC_{\sigma_c} \bD \br$, where~$\br$ is a sample from the standard normal. Intuitively, compared to the diagonal~$\bSigma_{z |\moving, \fixed}$ approximation, this sampling procedure smoothes the term~$\bD\br$ before adding the mean~$\bmu_{z|\moving,\fixed}$.

In our experiments, we show that this approximation yields smoother velocity fields during training, and the effect diminishes with higher~$\lambda$ values. However, the resulting deformation fields are diffeomorphic and accurate for \textit{both} approximations, demonstrating that the diagonal covariance approximation is sufficient when working with diffeomorphisms.

\begin{figure*}[tb!]
	\begin{center}
		\includegraphics[width=\linewidth]{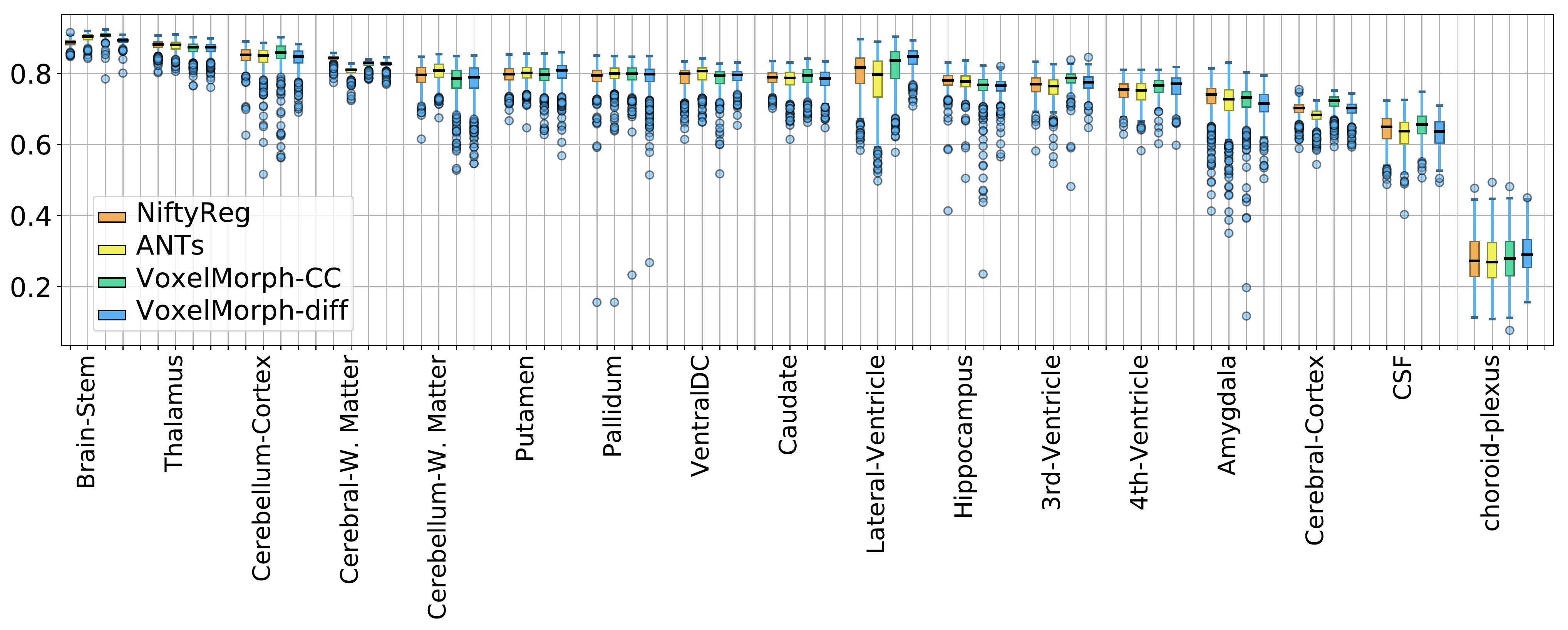}
	\end{center}
	\caption{Boxplots indicating Dice scores for anatomical structures for baselines ANTs, NiftiReg, VoxelMorph (CC), and finally our algorithm VoxelMorph-diff. Left and right hemisphere structures are merged for visualization, and ordered by average ANTs Dice score. In general, all four algorithms demonstrate comparable results, each performing slightly better in some structures and slightly worse in others. 
	}
	\label{fig:boxplot}
\end{figure*}

\section{Experiments}
\label{sec:results}


We perform a series of experiments demonstrating that the proposed probabilistic image registration framework achieves accuracy and runtime comparable to state-of-the-art methods while enabling diffeomorphic deformations. We also show the improvements enabled by the extended surface model, and analyze the effect of the various integration layers during test time. 

We focus on atlas-based registration, a common task in population analysis. Specifically, we register each scan to an atlas computed using external data~\citep{fischl2012,sridharan2013}. Because we implement our algorithm as part of the VoxelMorph framework, we will refer to it as \verb|VoxelMorph-diff|.

\begin{figure}[t!]
	\begin{center}
		\includegraphics[width=\linewidth]{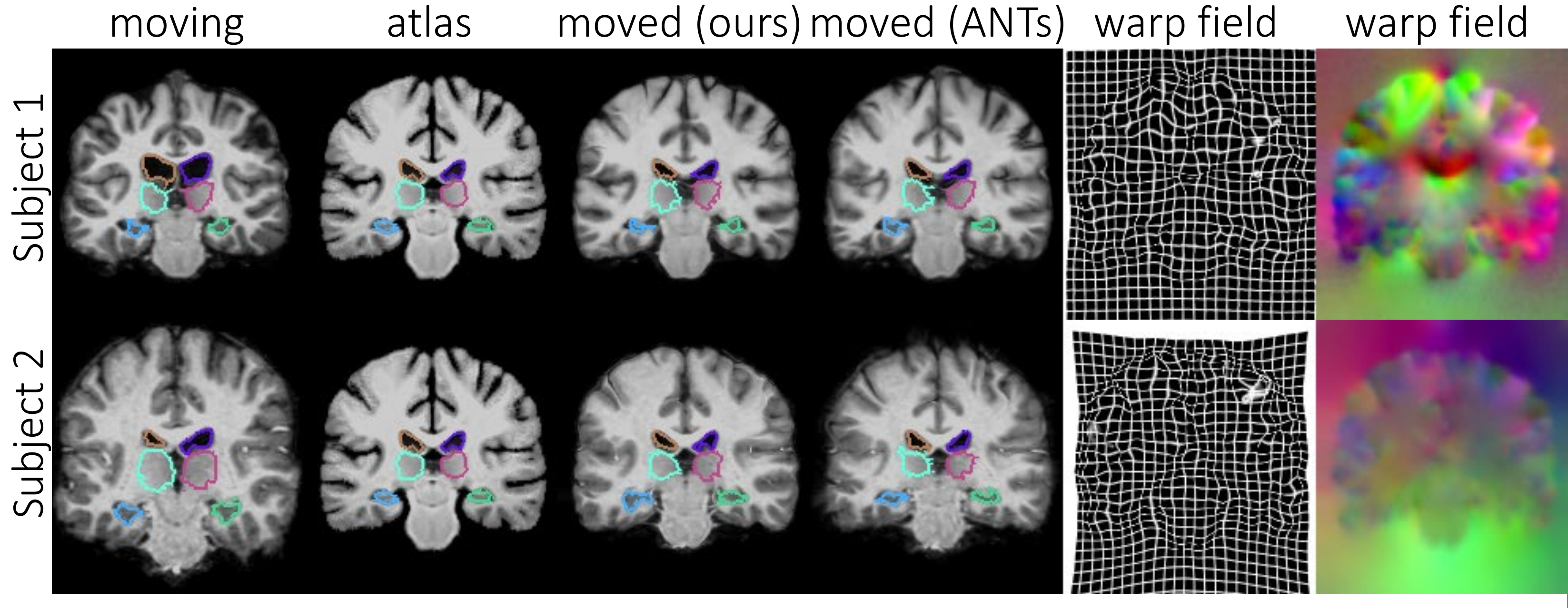}
	\end{center}
	\caption{Example MR slices of input moving image, atlas, and resulting warped image for our method and ANTs, with overlaid boundaries of ventricles, thalami and hippocampi. Our resulting registration field is shown as a warped grid and RGB image, {\color{blue}with the channels representing the x, y and z dimensions}. We omit VoxelMorph (CC) and NiftyReg examples, which are visually similar to our results and ANTs. More examples are provided in the supplementary material Figure~\ref{fig:sup:reg_examples}.
	}
	\label{fig:reg_examples}
\end{figure}

\begin{figure*}[t!]
	\centering
	\includegraphics[width=0.33\linewidth]{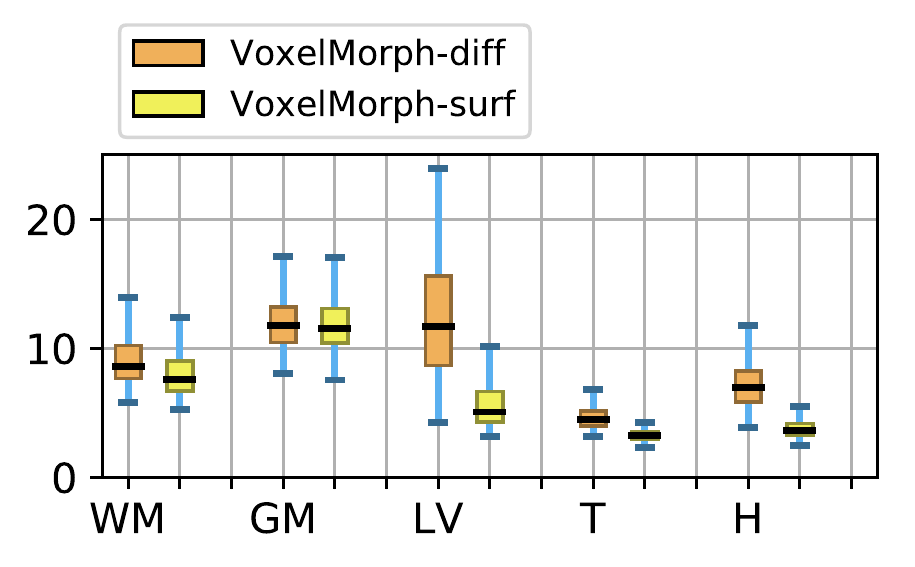}
\includegraphics[width=0.33\linewidth]{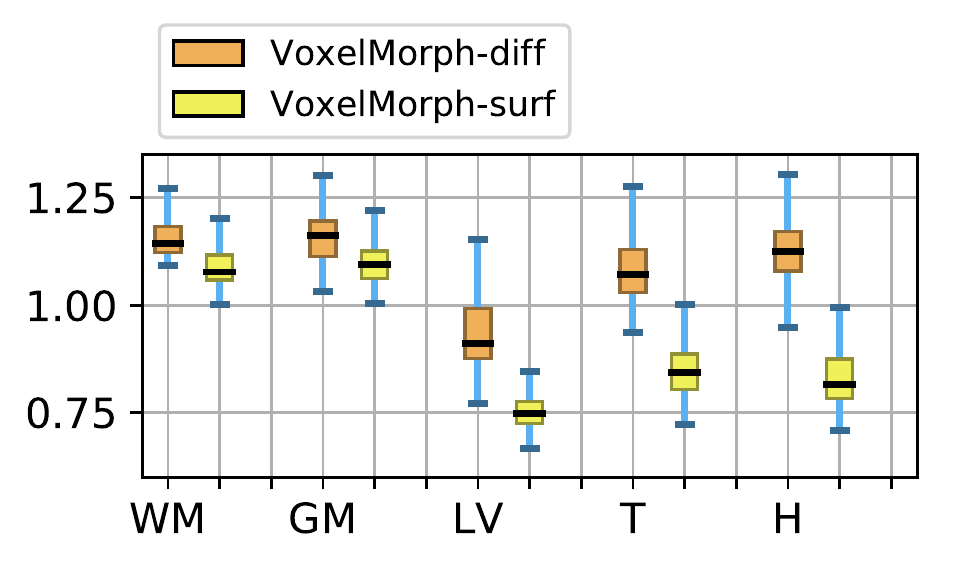}
\includegraphics[width=0.33\linewidth]{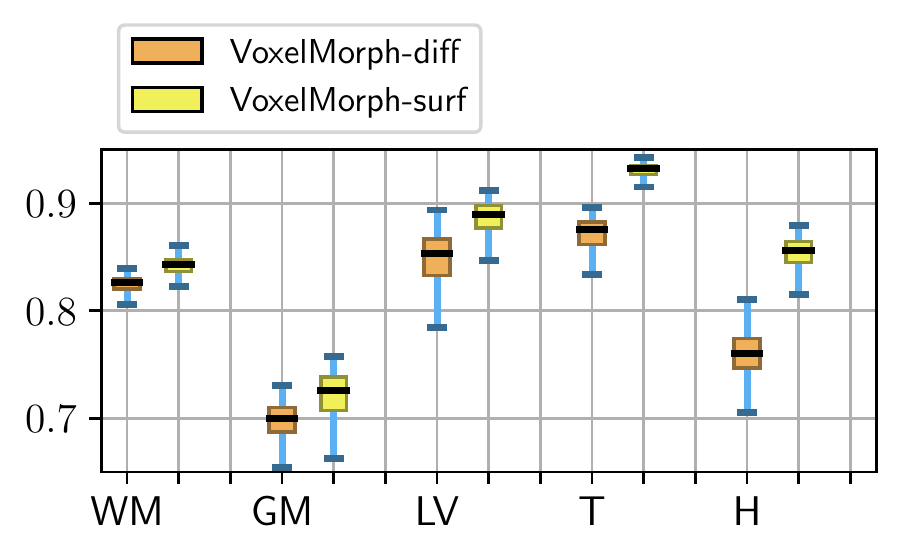}
	\caption{Surface results for the proposed VoxelMorph models. Left: maximum Euclidean surface distance (lower is better). Middle: median Euclidean surface distance (lower is better). Right: mean Dice (higher is better). VoxelMorph-surf trained with surfaces of the desired structures achieves significantly smaller surface distances and larger Dice scores on each structure.   We use left hemisphere white matter (WM), gray matter (GM), lateral ventricle (LV), Thalamus (T), and hippocampus (H).}
	\label{fig:surf-boxplot}
\end{figure*}

\subsection{Experiment setup}

\subsubsection{Data and Preprocessing}
We use a large-scale, multi-site dataset of 3731 T1-weighted brain MRI scans from eight publicly available datasets: OASIS~\citep{marcus2007open}, ABIDE~\citep{di2014autism}, ADHD200~\citep{milham2012adhd}, MCIC~\citep{gollub2013mcic}, PPMI~\citep{marek2011parkinson}, HABS~\citep{dagley2015harvard}, and Harvard GSP~\citep{holmes2015brain}. Acquisition details, subject age ranges and health conditions are different for each dataset. We performed standard pre-processing steps on all scans, including resampling to~$1$mm isotropic voxels, affine spatial normalization and brain extraction for each scan using FreeSurfer~\citep{fischl2012}. We crop the final images to $160\times 192 \times 224$. Segmentation maps including 29 anatomical structures, obtained using FreeSurfer for each scan, are used in evaluating registration results.  Each image contains roughly~$\sim1.6$ million brain voxels. We split the dataset into 3231, 250, and 250 volumes for train, validation, and test sets respectively, although we underscore that the training is unsupervised.

\subsubsection{Evaluation Metrics}
To evaluate a registration algorithm, we register each subject to an atlas, propagate the segmentation map using the resulting warp, and measure volume overlap using the Dice metric. For the surface experiments, we also employ the Euclidean surface distance, computed using the strategy described in~\eqref{eq:surface-approx}. 

We also evaluate the diffeomorphic property, a focus of our work. Specifically, the Jacobian matrix \mbox{$J_{\phi}(p) = \nabla \bphi (p) \in \mathcal{R}^{3\times 3}$} captures the local properties of $\bphi$ around voxel $p$. The local deformation is diffeomorphic, both invertible and orientation-preserving, only at locations for which $|J_{\phi}(p)| > 0$, where $| \cdot |$ is the determinant operator~\citep{ashburner2007}.
We count all other (folding) voxels, where~$|J_{\phi}(p)| \le 0$.

\subsubsection{Baseline Methods} We compare our approach with the popular ANTs software package using Symmetric Normalization (SyN)~\citep{avants2008}, a top-performing algorithm~\citep{klein2009}.  {\color{blue}We
	found that the default ANTs settings are sub-optimal for our task,
	and performed a wide parameter and similarity metric search
	across several datasets. We used the default geodesic implementation of SyN, which is most faithful to theoretical diffeomorphic development. Other versions, such as greedy SyN, would yield a slightly faster runtime, while giving less diffeomorphic deformations. We identified and use top performing
	parameter values for the Dice metric using: the cross-correlation
	(CC) loss function, SyN step size of 0.25, Gaussian smoothing
	of (9, 0.2) and three scales of 201 iterations.} We also test the NiftyReg package, for which we use a multi-threaded CPU implementation as a GPU implementation is not currently available.\footnote{We compiled the latest source code from March, 2018 (tree [4e4525]).} We experimented with different parameter settings for improved behavior, and used the following setting: CC cost function, grid spacing of 5, and 500 iterations.

To compare with recent learning-based registration approaches, we also test our recent CNN-based method, VoxelMorph, which produces state-of-the-art fast and accurate registration, but does not yield diffeomorphic results~\citep{balakrishnan2018a, balakrishnan2019}. We sweep the regularization parameter using our validation set, and use the optimal regularization parameter of 1 in our results.

{\color{blue} We also compute a supervised baseline by training a VoxelMorph-diff network using ground truth deformations. We build a ground truth dataset by registering over 650 atlas-MRI subject training pairs using NiftyReg with the described settings. We then train a neural network to predict the resulting deformation fields using a mean squared error (MSE) loss. We explored several variants, and found that doubling the model capacity by doubling the number of features at each layer, as well as penalizing the deformations fields only within the proximity of the atlas brain, yielded optimal results. To enable direct comparison, we used the \verb|VoxelMorph-diff| architecture, but without sampling of the velocity field.}


\subsection{Image Registration}

\begin{figure}[t!]
	\centering
	\includegraphics[width=0.9\linewidth]{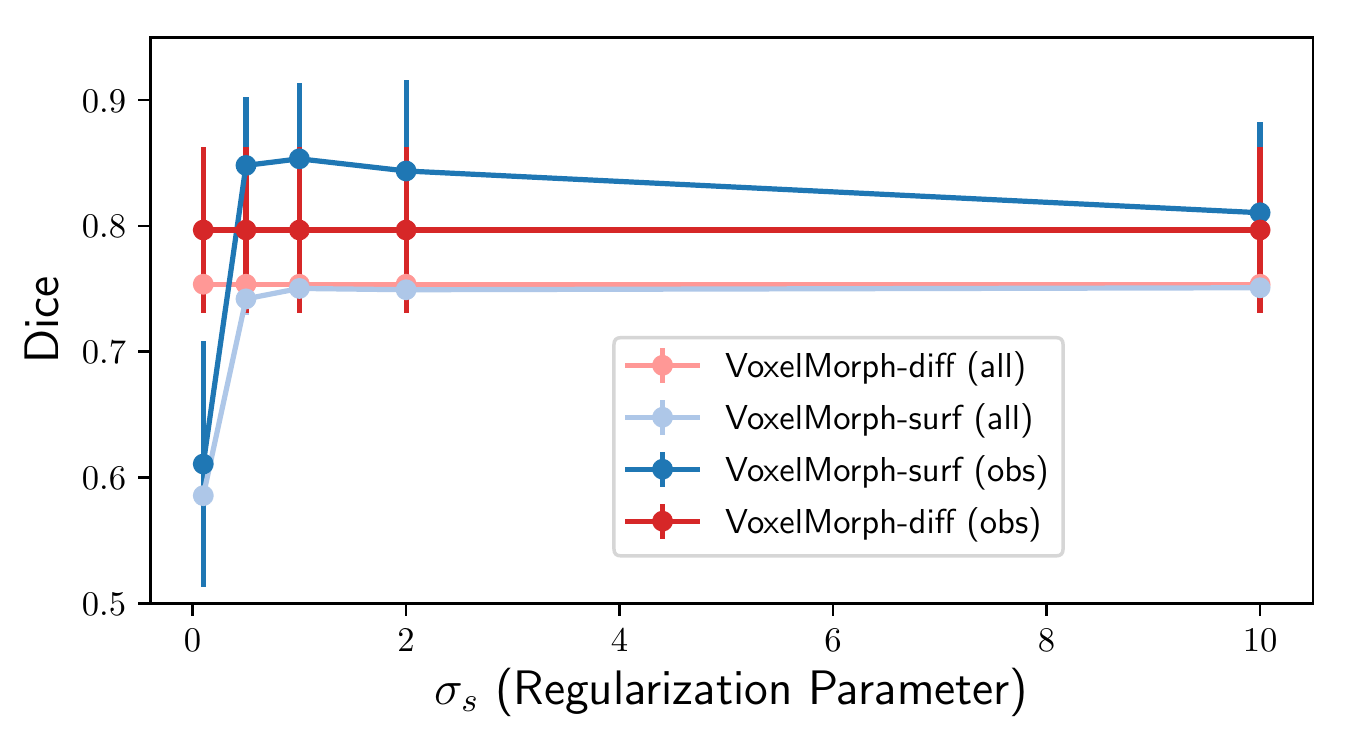}
	\caption{Average Dice score for VoxelMorph-surf models on the validation set. We test various values of the spatial noise parameter~$\sigma_s$, for both the desired structures observed  during training (obs) and all structures (all). For a range of values of~$\sigma_s\in[0.5, 2.0]$, we find significant increases for observed surfaces when using the generative surface model. }
	\label{fig:surf-hyperparams}
\end{figure}

\begin{figure}[h!]
	\begin{center}
		\includegraphics[width=\linewidth]{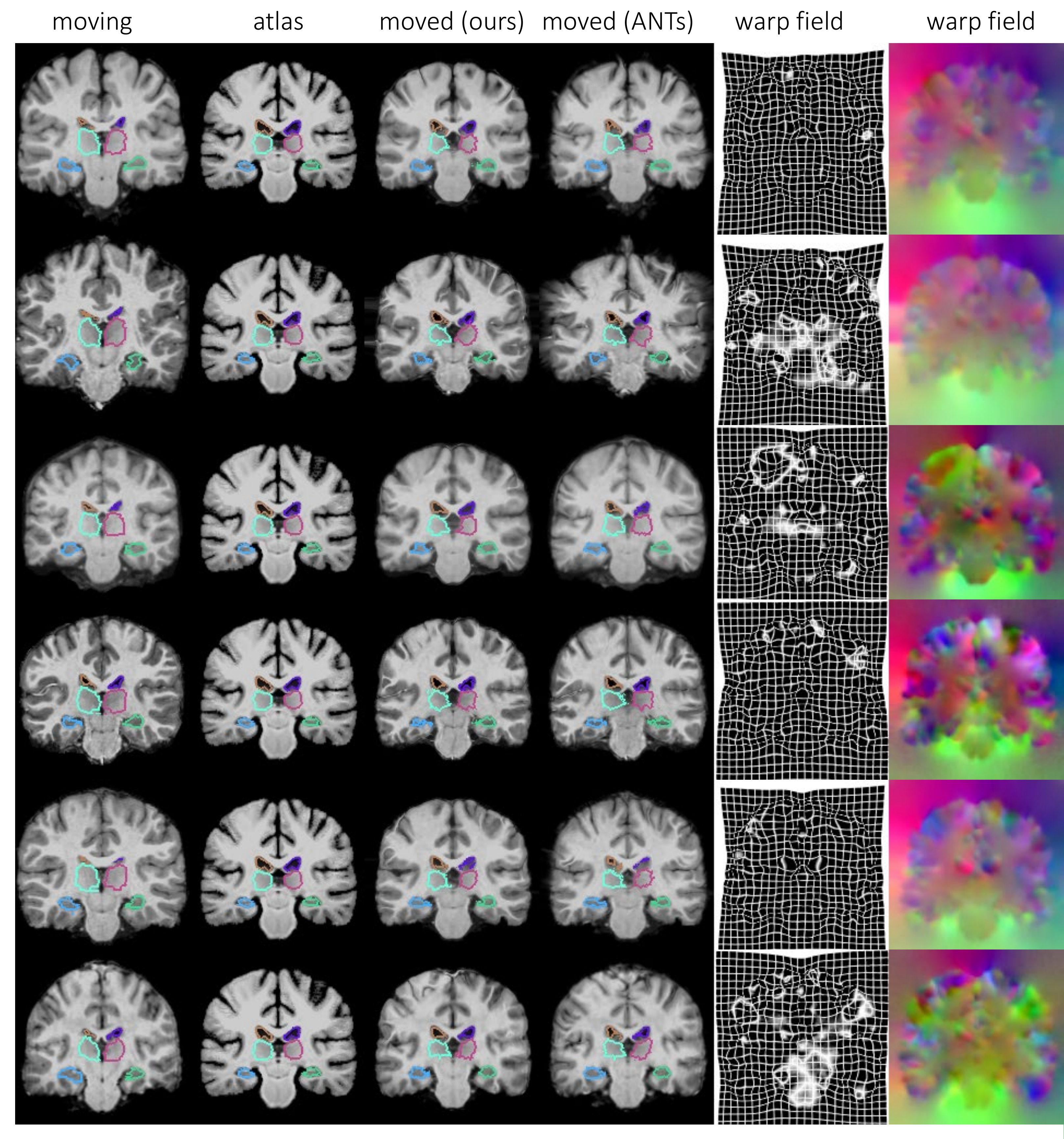}
	\end{center}
	\caption{Additional example MR slices of input moving image, atlas, and resulting warped image for our method (VoxelMorph-diff) and ANTs, with overlaid boundaries of ventricles, thalami and hippocampi. Each row is a different scan. Our resulting registration field is shown as a warped grid and RGB image, with each channel representing a dimension. We omit VoxelMorph and NiftyReg examples, which are visually similar to our results and ANTs. 
	}
	\label{fig:sup:reg_examples}
\end{figure}

Table~\ref{tbl:results} provides a summary of the results on the held-out test set. Figure~\ref{fig:reg_examples} and supplementary material Figure~\ref{fig:sup:reg_examples} show representative results. Figure~\ref{fig:boxplot} illustrates Dice results on several anatomical structures. For better visualization, we combine the same structures from the two hemispheres, such as the left and right hippocampus. Our algorithm, \verb|VoxelMorph-diff|, achieve state-of-the-art Dice results and runtimes, but produces diffeomorphic registration fields (nearly no folding voxels per scan) in a probabilistic framework. 

{\color{blue}All methods achieve comparable Dice results on each structure and overall, except the supervised method. Despite training the latter on 650 subjects, we found that the supervised network leads to more diffeomorphic deformations than the training deformations, but results in a slight loss in Dice score.} Learning-based methods require a fraction of the baseline runtimes to register two images: less than a second on a GPU, and less than a minute and a half on a CPU. Runtimes were computed for an NVIDIA TitanX GPU and a Intel Xeon (E5-2680) CPU, and exclude preprocessing common to all methods. 

Our method outputs positive Jacobians at nearly all voxels, which we analyze in more detail in a later section.{\color{blue} For \verb|VoxelMorph-diff|, we find that for most scans, the deformation fields result in zero folding voxels. Very few volumes lead to a few or tens of grouped folding	voxels, leading to a population average of less than a folding voxel per test scan.} {\color{blue} In contrast, the deformation fields resulting from the baseline methods contain a few thousand locations of non-positive Jacobians for each scan (Table~\ref{tbl:results}), usually grouped in clusters. This may be alleviated with increased spatial regularization or more optimization iterations, but this in turn leads to a drop in performance on the Dice metric or even longer runtimes. The table also shows that, at the presented settings, all methods result in an average Jacobian determinant close to~$1$, with \verb|VoxelMorph-diff| yielding smoothness statistics nearly identical to those given by NiftyReg, indicate smooth deformations.}

\begin{table}[tb!]
	\small
	\centering
	\begin{tabular}{c c c}
		\textbf{Method} & $|J| \le 0$ & $ \% \text{ of }|J| \le 0$\\
		\hline
		\hline
		\rule{0pt}{1.1em}
		ANTs SyN (CC) &9060 (4445)& 0.545 (0.267)\\
		NiftyReg (CC) &40425 (9901)& 2.431 (0.595)\\
		\algname{} (CC)&19077 (5928) & 1.147 (0.360)\\
		\hline
		\rule{0pt}{1.1em}
		\algname{}-diff                     & 0.1 (1.2) & 6.1e-6 (7.6e-5)\\
		\algname{}-surf (cer.w.m.)     & 3.0 (6.2)&  1.8e-4 (3.8e-4)\\
		\algname{}-surf (cer.cor.)   & 3.4 (6.4) & 2.0e-4 (3.9e-4)\\
		\algname{}-surf (lat.ven.)     & 4.0 (8.0) & 2.3e-4 (4.8e-4)\\
		\algname{}-surf (thalamus)          & 4.3 (8.2) & 2.6e-4 (4.9e-4)\\
		\algname{}-surf (hip.)       & 2.7 (5.8) & 1.6e-4 (3.5e-4)\\
		\hline
	\end{tabular}
	\vspace{0.1cm}
	\caption{Regularity measures for image and surface models on the test set. Leveraging diffeomorphic aspect of our joint image and surface model, VoxelMorph-surf preserved very low numbers of folding voxels even when training with example surfaces.}
	\label{tbl:jacobian-results}
\end{table}

\subsection{Image and Surface Registration}

	In this section, we evaluate the generative surface model. We demonstrate the use of anatomical segmentation alongside images during training, and refer to this model as~\verb|VoxelMorph-Surf|. We focus on the setting where one structure of interest is available during training, and learn separate networks for the left white matter, gray matter, ventricle, thalamus and hippocampus. Our goal is to analyze how the additional surface model terms affect the accuracy and regularity of resulting deformations.

\begin{figure*}[t!]
	\centering
	\includegraphics[width=0.49\linewidth]{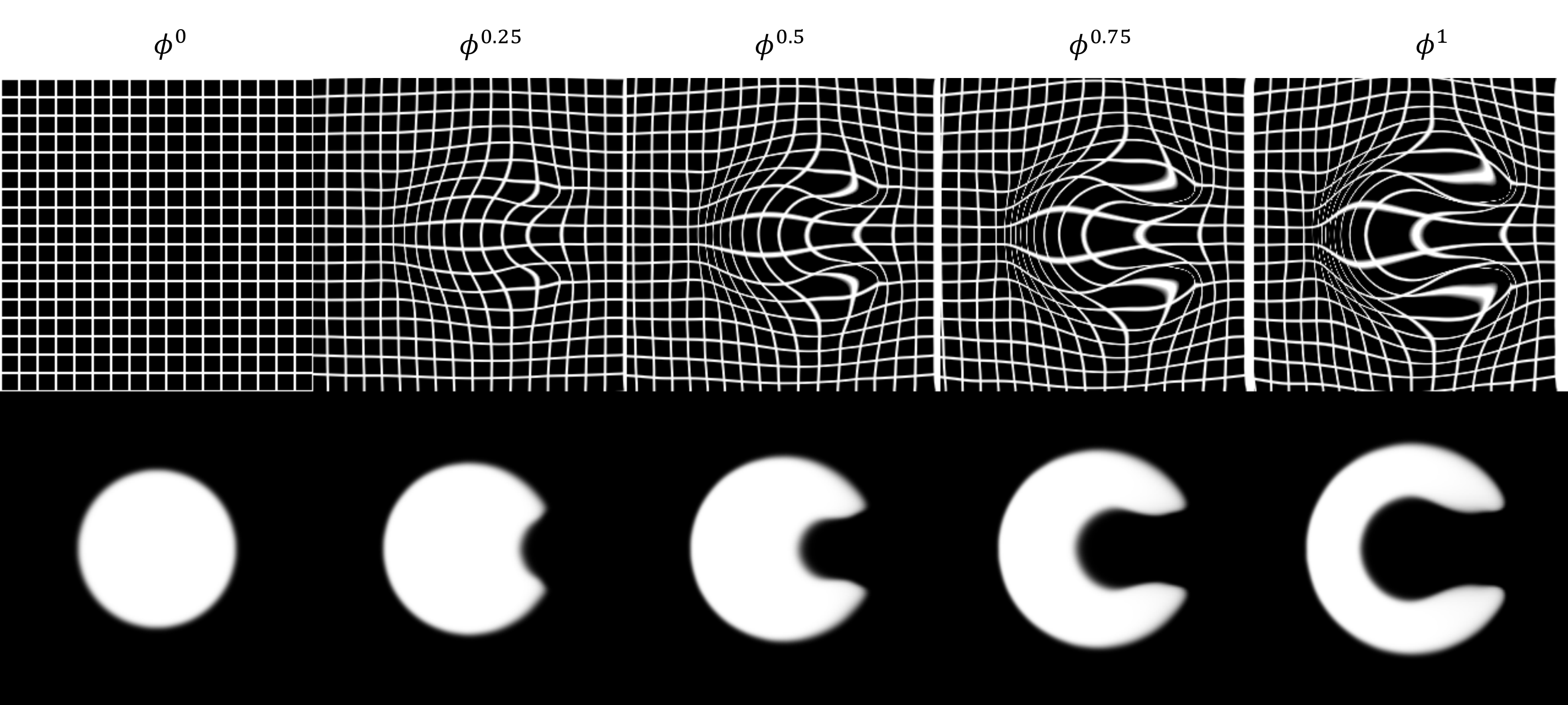}
	\vspace{0.3cm}
	\includegraphics[width=0.49\linewidth]{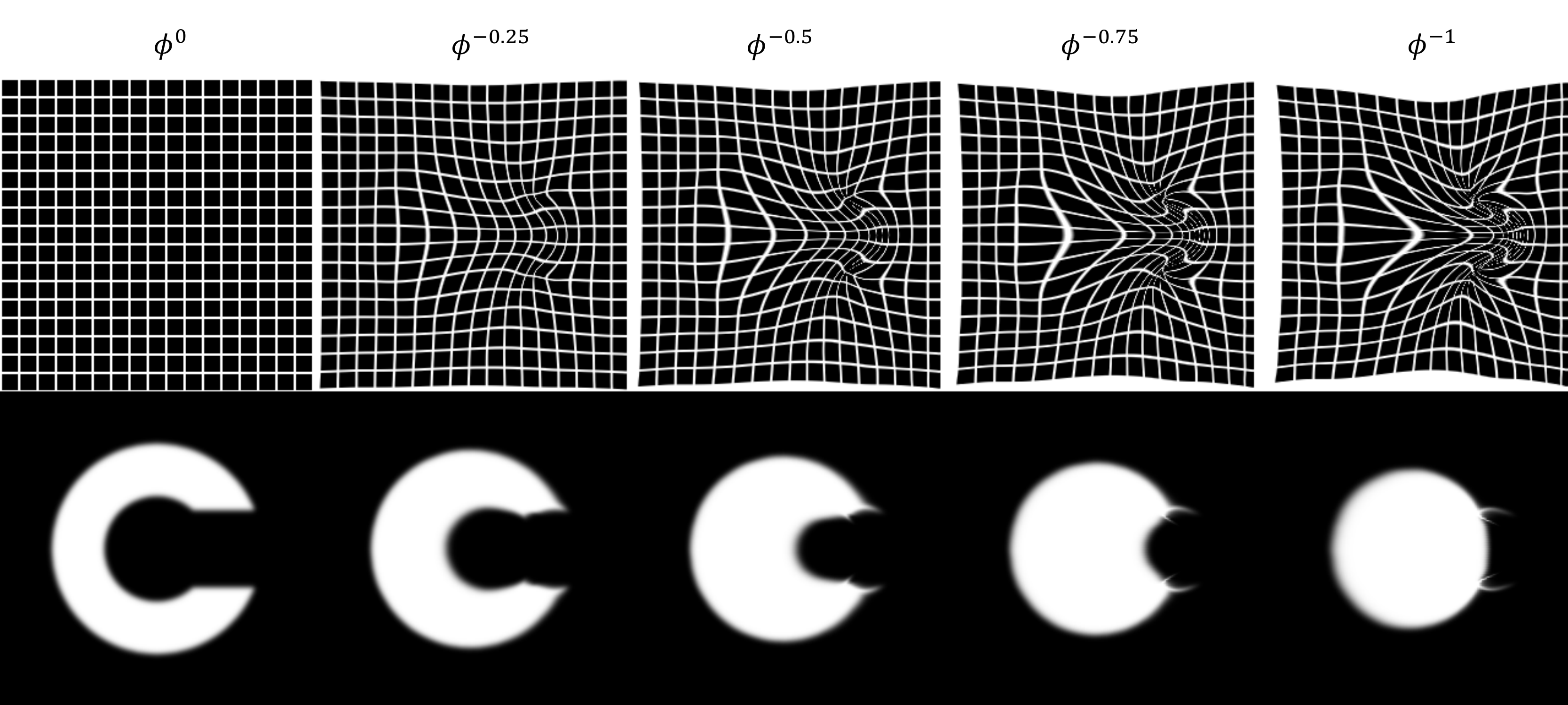}
	\includegraphics[width=1\linewidth]{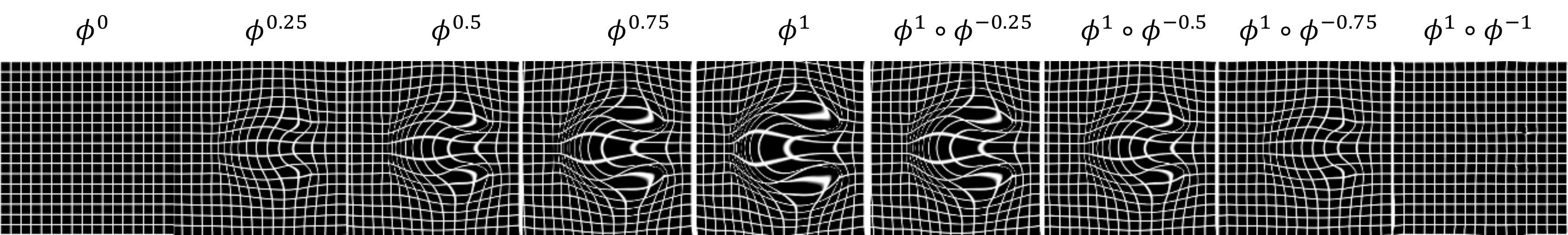}
	\hfill
	\caption{\color{blue} C-shape controlled experiments. We learn to warp disks to Cs of different radii, and illustrate the registration results for one example. The top row illustrates the integration of the velocity field at different time points, and the second row shows the resulting warp of the circle or C. Finally, on the bottom row, we illustrate deforming the grid with a composition of the forward warp and the inverse warp, demonstrating a return to identity.}
	\label{fig:C_shape}
\end{figure*}

	Figure~\ref{fig:surf-hyperparams} illustrates the behaviour of the model with respect to the hyper-parameter~$\sigma_s$ on the validation set. For a range of~$\sigma_s$ values, we find a significant improvement in terms of Dice for the desired structure. For very small values of~$\sigma_s$, the training becomes unstable leading to poor generalization. A very large~$\sigma_s$ value leads to the model ignoring the surface term. Since the Dice scores are comparable in the range~$\sigma_c \in [0.5, 2]$, for the rest of this section we use~$\sigma_s = 2$, which exhibits slightly fewer folding voxels ($\le5$ compared to $\sim20$ for $\sigma_c=1$).
	
	Figure~\ref{fig:surf-boxplot} demonstrates the improvement on the test set in terms of Euclidean surface distance and Dice, compared to the image-only registration model \verb|VoxelMorph-diff|. \verb|VoxelMorph-surf| improves significantly in all measures for most desired structures. Additionally, Table~\ref{tbl:jacobian-results} illustrates that with increased accuracy in both metrics, the number of folding voxels in the entire volume increases only very slightly (to an average 3.5 voxels per volume), which remains orders of magnitude fewer than the baseline methods (Table~\ref{tbl:results}). Figure~\ref{fig:sup:reg_examples_supervised} in the supplementary material illustrates example results.

	In summary, the principled joint diffeomorphic model enables the use of surfaces during training which dramatically improves registration near a given structure while preserving desired deformation properties. For example, given hippocampus surfaces at training, registration using \verb|VoxelMorph-Surf| improves Dice by~$\sim 9$ points over \verb|VoxelMorph-diff|, improves maximum surface distance by more than three voxels, and preserving diffeomorphisms (less than three folding voxels per scan).

\begin{figure}[t!]
	\centering
	\includegraphics[width=0.9\linewidth]{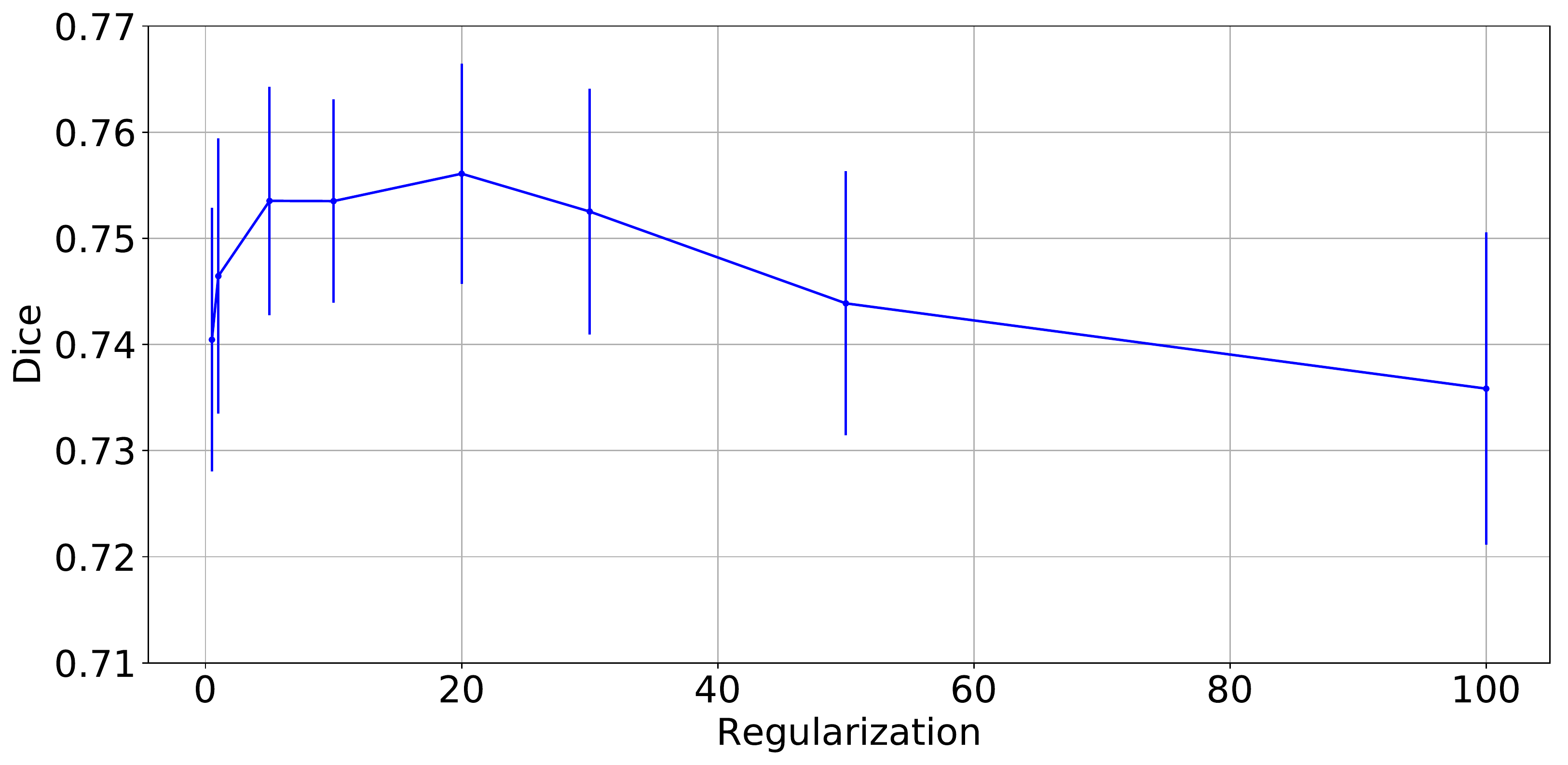}
	\caption{Dice score (computed using 50 validation scans) for VoxelMorph-diff with various values of the precision parameter~$\lambda$.}
	\label{fig:sweep}
\end{figure}


\subsection{Analysis}

\subsubsection{Parameter Analysis} 
The two main hyper-parameters, smoothing precision~$\lambda$ and image noise~$\sigma_I^2$, have physical meaning in our generative model. However, they share a single degree of freedom in the loss function.  We set~$\sigma_I^2 = 0.02$, and vary the precision scale~$\lambda$ between 0.5 and 100. Figure~\ref{fig:sweep} shows average Dice scores for 50 validation set scans for different parameter values, showing that the results vary smoothly over a large range, with reasonable behavior even near~$\lambda\sim 0$. We use~$\lambda=20$ in our experiments above. 

%


{\color{blue} 
	\subsubsection{C-shape Registration}
	We also perform analysis on controlled experiments with C-shape synthetic images with intensities in~$[0,1]$. Specifically, we train a \verb|VoxelMorph-diff| network to learn to register a disk with a radius ranging from one third to one fifth of the image, to a C-shape with variable radius and thickness. The outer radius of the C shape is sampled uniformly in the range~$[1/3.5, 1/2.5]$ of the image size, whereas the inner radius is in the range~$[1/6.5, 1/5.5]$. We increase hyper-parameter~$\sigma_s = 0.06$ to account for the increase in maximum intensity. Figure~\ref{fig:C_shape} illustrates representative images and deformation fields. To obtain the fields at intermediate time points between 0 and 1, we employ Tensorflow ODE solver. We find that all the deformation fields lead to accurate registration between disks and C shapes, and have no folding voxels. We also find that the deformation fields are invertible, bringing the grid back to identity when the transforms are composed. }

\begin{figure}[t!]
	\begin{center}
		\includegraphics[width=1\linewidth]{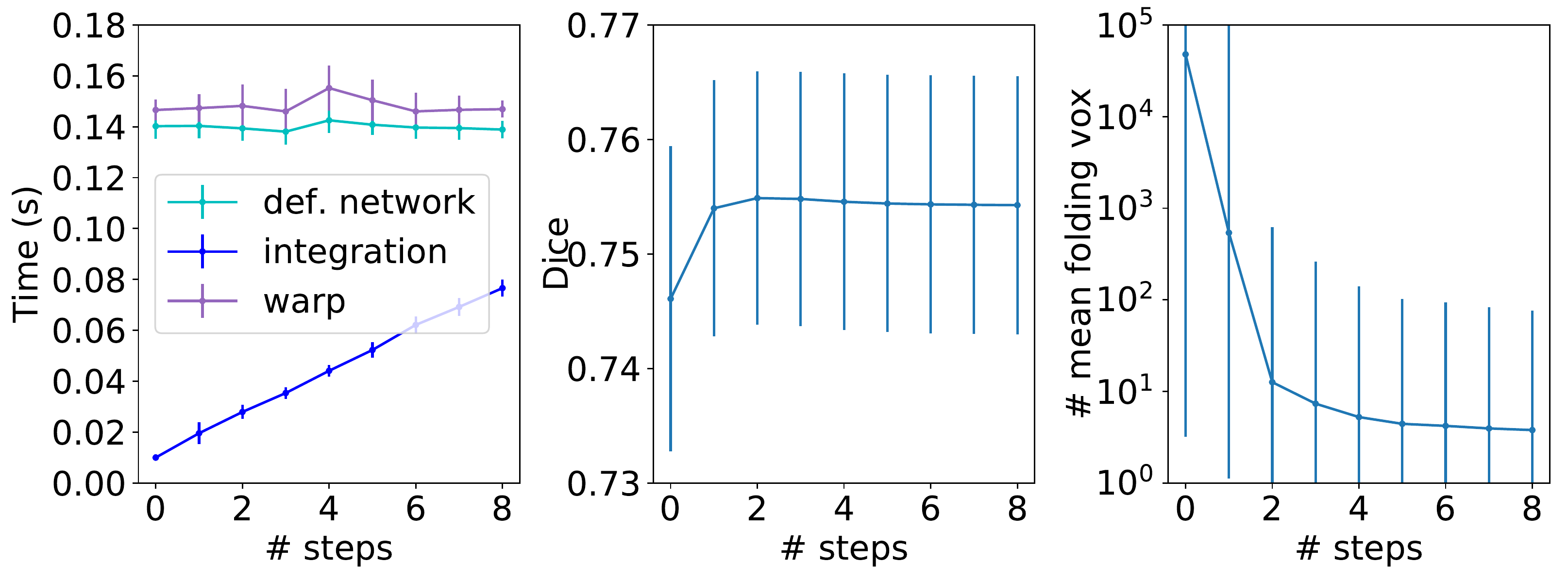}
		\includegraphics[width=1\linewidth]{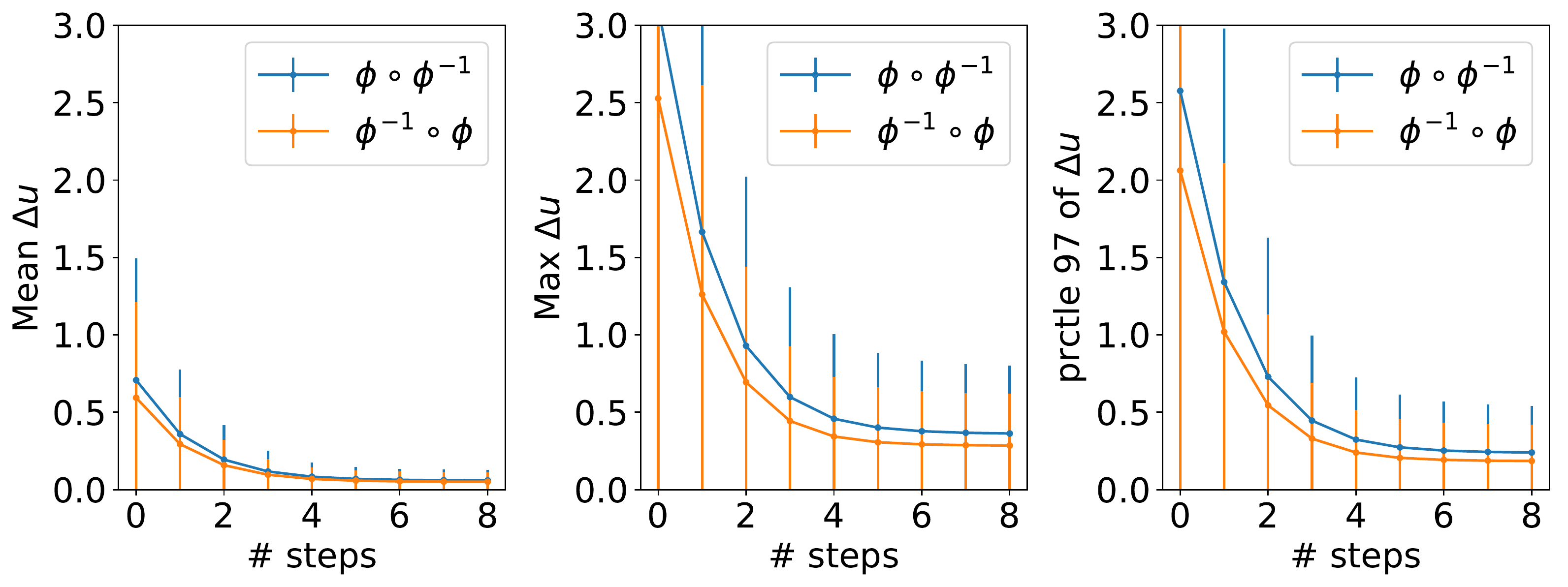}
	\end{center}
	\caption{The effect of different number of scaling and squaring steps on the registration accuracy, runtime, deformation regularity and invertability. We find that after five scaling and squaring steps, our model, VoxelMorph-diff, is able to produce state-of-the-art accuracy while having essentially no folding (note the log-scale vertical axis in the top-right graph). Similarly, it is able to produce invertible deformations, as seen by the measure of the displacement error~$\Delta \bu = |I - \phi \circ \phi^{-1}|$. The total runtime cost of the scaling and squaring operations is below the runtime of the rest of the networks, indicating that increasing the number of steps improves deformation properties for trivial runtime cost.}
	\label{fig:expt_integration_steps}
\end{figure}

\subsubsection{Integration Steps}

	During training, we hold the number of \textit{scaling and squaring} steps fixed. However, this number can be varied at test time, affecting aspects of the resulting deformation field. In this section, we analyze the effects of the number of steps on accuracy, runtime, field regularity and invertability. We perform this experiment using 50 validation subjects and the image registration network VoxelMorph-diff trained with $T=7$ integration steps and regularization parameter~$\lambda=20$. The velocity field is computed every two voxels, but all of the conclusions in this section are likely to apply to many reasonable field spacings.
	
	Figure~\ref{fig:expt_integration_steps} summarizes the analysis results. The runtime increases modestly with the number of steps, and is overall significantly smaller than the cost of the rest of the network (i.e.\ the deformation network computation of the velocity field, and the spatial transform of the full moving image). After four scaling and squaring steps, the method achieved maximum Dice score. We observe a steep decline in the number of folding voxels (note the log-scale vertical axis), reaching less than five voxels after five scaling and squaring steps, compared to classical methods which can include thousands of such voxels (Table~\ref{tbl:results}). Finally, we measure the average displacement error after inverting the deformation fields:~$\Delta \bu = |Id - \bphi \circ \bphi^{-1}|$. We find that after five scaling and squaring steps, even the \textit{worst} error is under a half voxel, indicating that five steps are sufficient to ensure invertible deformations.
	
{\color{blue} In addition, we implemented the integration of the velocity field using Tensorflow ODE solvers and using standard quadrature,	but found that these required significantly more runtime
	compared to the scaling and squaring strategy, consistent with 
	literature findings~\citep{arsigny2006,ashburner2007,modat2014global}.}  Specifically, while five scaling and squaring operations required~$0.06\pm0.01$ seconds, equivalent quadrature integration required~$64$ operations (occupying prohibitive amounts of memory) and~$0.53\pm0.01$ seconds, and ODE-solver based integration with default parameter required a single layer and~$2.9 \pm 0.1$ seconds. At comparable integration settings such as these, all three methods achieve similar Dice scores of $0.75 \pm 0.01$. While these alternative methods require significant resources, all three implementations are available in our source code for experimentation.
	
	This analysis indicates that the proposed scaling and squaring network integration layer is efficient and accurate. Increasing the number of scaling and squaring layers incurs a negligible runtime cost while improving deformation field properties. We use~$T=7$ squaring steps in the test experiments above.

\begin{figure}[t!]
	\includegraphics[width=1\linewidth]{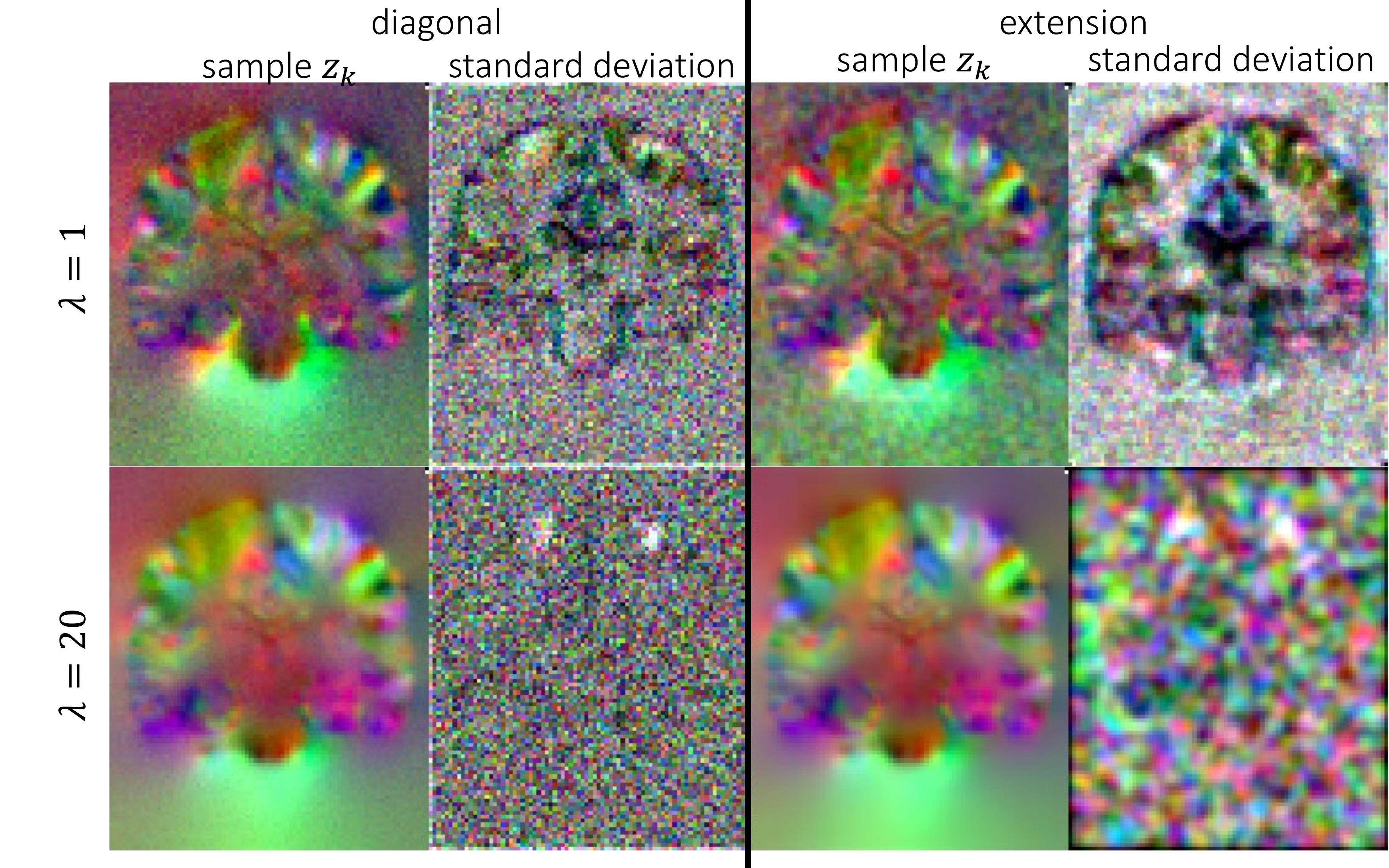}
	\caption{Illustration of voxel independence assumption in variational approximations for two prior parameters~$\lambda=1$ (top) and~$\lambda=20$ (bottom). Each row contains an example sample velocity field~$\bz_k$, and the voxel-wise standard deviation over 500 samples for that subject.}
	\label{fig:smooth_approximation}
\end{figure}


\begin{table}[!t]
		\vspace{-0.1cm}
		\small
	\begin{tabular}{c | c  c | c c}
		& \multicolumn{2}{c|}{Dice}   & \multicolumn{2}{c}{$|J| \le 0$} \\
		\hline
		$\lambda$ & diagonal  & extension & diagonal  & extension \\
		\hline
		$1$ & 0.74 (0.01) & 0.74 (0.01) & 2934 (2007) & 2720 (1593) \\
		$20$ & 0.75 (0.01) & 0.75 (0.01) & 0.32 (0.96) & 0.16 (0.57) \\
		\hline
	\end{tabular}
	\vspace{0.1cm}
	\caption{Accuracy and deformation regularity for the different variational approximations and two dramatically different values for smoothing parameter~$\lambda$. We find that for a given parameter value, the approximations lead to comparable accuracy and number of folding voxels.}
	\label{tab:smooth_approximation}
\end{table}

\vfill

\subsubsection{Velocity Sampling and Uncertainty}

	We also evaluate the modeling assumptions of the variational covariance~$\bSigma_{z |\moving, \fixed}$. Figure~\ref{fig:smooth_approximation} illustrates example samples of the velocity field~$\bz_k$ and voxel-wise empirical variance for the two~$\bSigma_{z | \moving, \fixed}$ approximations: diagonal covariance and the extended approximation in Section~\ref{sec:nondiag-approx} that smooths samples~$\bz_k$. For under-regularized networks (very low values for hyper-parameter~$\lambda$), the latter approximation yields smoother velocity fields. However, given a higher hyper-parameter~$\lambda$ value, such as the one used in our experiments, the network learns smaller values for the diagonal~$\bSigma_{z | \moving, \fixed}$ approximation, and yields smooth samples~$\bz_k$ with either method. Futhermore, despite the difference in smoothness of the velocity field samples~$\bz_k$, the integration operation leads to equally regular and accurate deformations~$\bphi_{z_k}$ for a given~$\lambda$ (Table~\ref{tab:smooth_approximation}). 
	
	Therefore, although the diagonal covariance has the potential to add noise to velocity field samples, the loss function coupled with the integration operation lead to smooth and accurate deformation fields~$\bphi_z$ at reasonable~$\lambda$ values.  {\color{blue}Therefore, in the current setting, the diagonal and non-diagonal covariances give similar results.
	Nonetheless, in other applications the non-diagonal covariance might be important. For example,
	diagonal covariances would likely have negative effects in a different deformation model, for instance if~$z$ was modelled as the displacement field itself.}

\vspace{0.2cm}
\section{Discussion and Conclusion}

In this work, we build a principled connection between classical registration methods and recent learning-based approaches. We propose a probabilistic model for diffeomorphic image registration and derive a learning algorithm that leverages a convolutional neural network and unsupervised, end-to-end learning for fast runtime. To achieve diffeomorphic transforms, we integrate stationary velocity fields through novel \textit{scaling and squaring} differentiable network operations, and provide implementation and analysis for other integration layers. 

Although the simplifying diagonal approximation to the velocity covariance~$\bSigma_{z |\moving, \fixed}$ adds voxel-independent noise to every velocity field sample~$\bz_k$, the resulting deformation fields are well behaved because of our smoothing prior and diffeomorphic representation.

We also provide an anatomical surface deformation model. If image segmentations are available for a particular anatomical structure, the generative model incorporates them naturally in the same joint framework during training, while not requiring the surfaces at test time.

Our algorithm can infer the registration of new image pairs in under a second. Compared to traditional methods, our approach is significantly faster, and compared to recent learning based methods, our method offers diffeomorphic guarantees. We demonstrate that the surface extension to our model can help improve registration while preserving properties such as low runtime and diffeomorphisms.

{\color{blue} Furthermore, several conclusions shown in recent papers apply to our method. For example, when only given very limited training data, deformation from VoxelMorph can still be used as initialization to a classical method, enabling faster convergence (Balakrishnan et al, 2019).}

Our focus in this framework has been to present the technical connection between classical and learning paradigms, and show that diffeomorphisms are attainable in a very low runtime. Immediate extensions can enable other models and applications. For example, our derivation is generalizable to other formulations:~$\bz$ can be a low dimensional embedding representation of a deformation field, or the displacement field itself. Similarly, the variational covariance~$\bSigma_{z|\moving,\fixed}$ enables an estimation of the uncertainty of the deformation field at each voxel, which can be informative in downstream tasks such as biomedical segmentation or population analysis. The model is also widely applicable to other applications, such as subject-to-subject registration, segmentation-only registration, or using multiple surfaces to improve image-based registration.

\vspace{-0.15cm}
\section{Acknowledgments}
\vspace{-0.1cm}
This research was funded by NIH grants R01LM012719, R01AG053949, and 1R21AG050122, NSF CAREER 1748377, NSF NeuroNex Grant 1707312, and Wistron Corporation. {\color{blue} We thank John Ashburner for sharing code to simulate images for the C-shape controlled experiments.}

\bibliographystyle{plain}
\bibliography{references}

\clearpage
\onecolumn
\section*{Supplementary Material}

\subsection*{Derivation of Main Loss}
\label{sec:sup:loss-derivation}

\begin{align}
\mathcal{L}(\bpsi; \bmoving, \bfixed)  &= - \Expect_{q} \left[ \log p(\bmoving | \bz ; \bfixed) \right] + \KL \left[  q_{\bpsi}(\bz|\bmoving; \bfixed) ||  p(\bz)  \right]  \nonumber \\
&= - \Expect_{q} \left[ \log \mathcal{N}(\bmoving ; \bz \circ \bfixed; \sigma^2 \mathbb{I}) \right] 
+ \KL \left[  \mathcal{N}(\bz; \bmu_{z | \moving, \fixed}, \bSigma_{z | \moving, \fixed}) ||  \mathcal{N}(\bz; \b0, \bLambda_z)  \right]  \nonumber \\
&= \frac{1}{2} \Expect_{q} \left[ \log 2\pi \sigma^{2d} + \frac{1}{\sigma^2}\|\bmoving - \bfixed \circ \bphi_{z}\|^2
\right] 
+ \KL \left[  \mathcal{N}(\bz; \bmu_{z | \moving, \fixed}, \bSigma_{z | \moving, \fixed}) ||  \mathcal{N}(\bz; \b0, \bLambda_z)  \right]  \nonumber \\
&= \frac{1}{2\sigma^2}  \Expect_{q} \left[ \|\bmoving - \bfixed \circ \bphi_{z_k}\|^2 \right]
+ \frac{1}{2} \left[ \log\frac{|\bLambda_z^{-1}|}{|\bSigma_{z | \moving, \fixed}|} - 3d + \text{tr}(\bLambda_z \bSigma_{z | \moving, \fixed})  + \bmu_{z | \moving, \fixed}^T \bLambda_z \bmu_{z | \moving, \fixed} \right] + \text{const}. \nonumber
\end{align}
Using the facts that~$\log |\bLambda_z|$ is constant,~$\log|\bSigma_{z | \moving, \fixed}| = \text{tr}\log\bSigma_{z | \moving, \fixed}$, and~$ \text{tr}(\bLambda_z \bSigma_{z | \moving, \fixed}) = \text{tr}((\lambda \bD - \bA) \bSigma_{z | \moving, \fixed}) = \text{tr}(\lambda \bD \bSigma_{z | \moving, \fixed})$, and approximating the expectation with~$K$ samples~$z_k \sim q_z$, we obtain
\begin{align}
\mathcal{L}(\bpsi; \bmoving, \bfixed) &= \frac{1}{2\sigma^2K} \sum_k \|\bmoving - \bfixed \circ \bphi_{z_k}\|^2 + \frac{1}{2} \left[ \text{tr}(\lambda\bD \bSigma_{z|x;y} - \log\bSigma_{z|x;y}) + \bmu_{z | \moving, \fixed}^T \bLambda_z \bmu_{z | \moving, \fixed} \right] + \text{const}.
\label{eq:sup:main_loss}
\end{align}

\subsection*{Derivation of Surface VLB and Loss}
\label{sec:sup:surf-derivation}

We derive the variational lower bound and loss for the generative surface model. We start by minimizing the KL divergence between the true and approximate posteriors:
\begin{align}
&\min_{\psi} \KL \left[q_{\bpsi}(\bz|\bfixed ; \bmoving) || p(\bz|\bfixed, \bs_{\bfixed}; \bmoving, \bs_{\bmoving}) \right] \nonumber \\
&= \min_{\psi} \Expect_{q} \left[ \log q_{\bpsi}(\bz|\bfixed ; \bmoving) - \log p(\bz|\bfixed, \bs_{\bfixed}; \bmoving, \bs_{\bmoving}) \right] \nonumber \\
&= \min_{\psi} \Expect_{q} \left[ \log q_{\bpsi}(\bz|\bfixed ; \bmoving) - \log p(\bz, \bfixed, \bs_{\bfixed}; \bmoving, \bs_{\bmoving}) \right] + \log p(\bfixed, \bs_{\bfixed}; \bmoving, \bs_{\bmoving}) \nonumber \\
&= \min_{\psi} \Expect_{q} \left[ \log q_{\bpsi}(\bz|\bfixed ; \bmoving) - \log p(\bz) - \log p(\bfixed, \bs_{\bfixed} | \bz ; \bmoving, \bs_{\bmoving}) \right] + \text{const} \nonumber \\
&\stackrel{\star}{=} \min_{\psi} \Expect_{q} \left[ \log q_{\bpsi}(\bz|\bfixed ; \bmoving) - \log p(\bz) - \log p(\bfixed | \bz; \bmoving) - \log p(\bs_{\bfixed} | \bz ; \bs_{\bmoving}) \right] + \text{const} \nonumber \\
&= \min_{\psi} \KL \left[  q_{\bpsi}(\bz|\bfixed; \bmoving) ||  p(\bz)  \right] - \Expect_{q} \left[ \log p(\bfixed | \bz ; \bmoving) \right] -  \Expect_{q} \left[ \log p(\bs_{\bfixed} | \bz ; \bs_{\bmoving}) \right] + \text{const},
\label{eq:sup:surf-VLB}
\end{align}
where in~$\star$ we used the assumptions that the fixed image is independent of anatomical surfaces given the moving image and the deformation, and the fixed surface is independent of either image given the moving surface and the deformation. Following this variational lower bound, the loss follows the previous section closely (see~\eqref{eq:sup:surf-VLB}), with the additional term:
\begin{align}
\Expect_{q} \left[ \log p(\bs_{\bfixed} | \bz ; \bs_{\bmoving}) \right] &= \Expect_{q} \left[ \log \mathcal{N}(\bmoving ; \bz \circ \bfixed; \sigma^2 \mathbb{I}) \right] \nonumber \\
&= \frac{1}{2} \Expect_{q} \left[\log 2\pi \sigma_s^{2d} + \frac{1}{\sigma_s^2} \|\bs_{\bfixed} - \bs_{\bmoving} \circ \bphi_{z_k}\|^2 \right] \nonumber \\
&= \frac{1}{2} \Expect_{q} \left[\frac{1}{\sigma_s^2} \|\bs_{\bfixed} - \bs_{\bmoving} \circ \bphi_{z_k}\|^2 \right] + \text{const.}
\end{align}
Combining this term with~\eqref{eq:sup:main_loss}, and approximating expectations with~$k$ samples leads to the final loss~\eqref{eq:surf-loss}:
\begin{align}
\mathcal{L}(\bpsi; \bfixed, \bs_{\bfixed}, \bmoving, \bs_{\bmoving})  
&= \frac{1}{2\sigma^2K} \sum_k ||\bfixed - \bmoving \circ \bphi_{z_k}||^2  + \frac{1}{2\sigma_s^2K} \sum_k ||\bs_{\bfixed} - \bs_{\bmoving} \circ \bphi_{z_k}||^2 \nonumber \\
&+ \frac{1}{2} \left[ \text{tr}(\lambda\bD \bSigma_{z|x;y} - \log\bSigma_{z|x;y}) + \bmu_{z | \moving, \fixed}^T \bLambda_z \bmu_{z | \moving, \fixed} \right] + \text{const}.
\end{align}

\clearpage
\section*{Overview figure with surface loss}

\begin{figure*}[h!]
	\centering
	\begin{center}
		\includegraphics[width=1\linewidth]{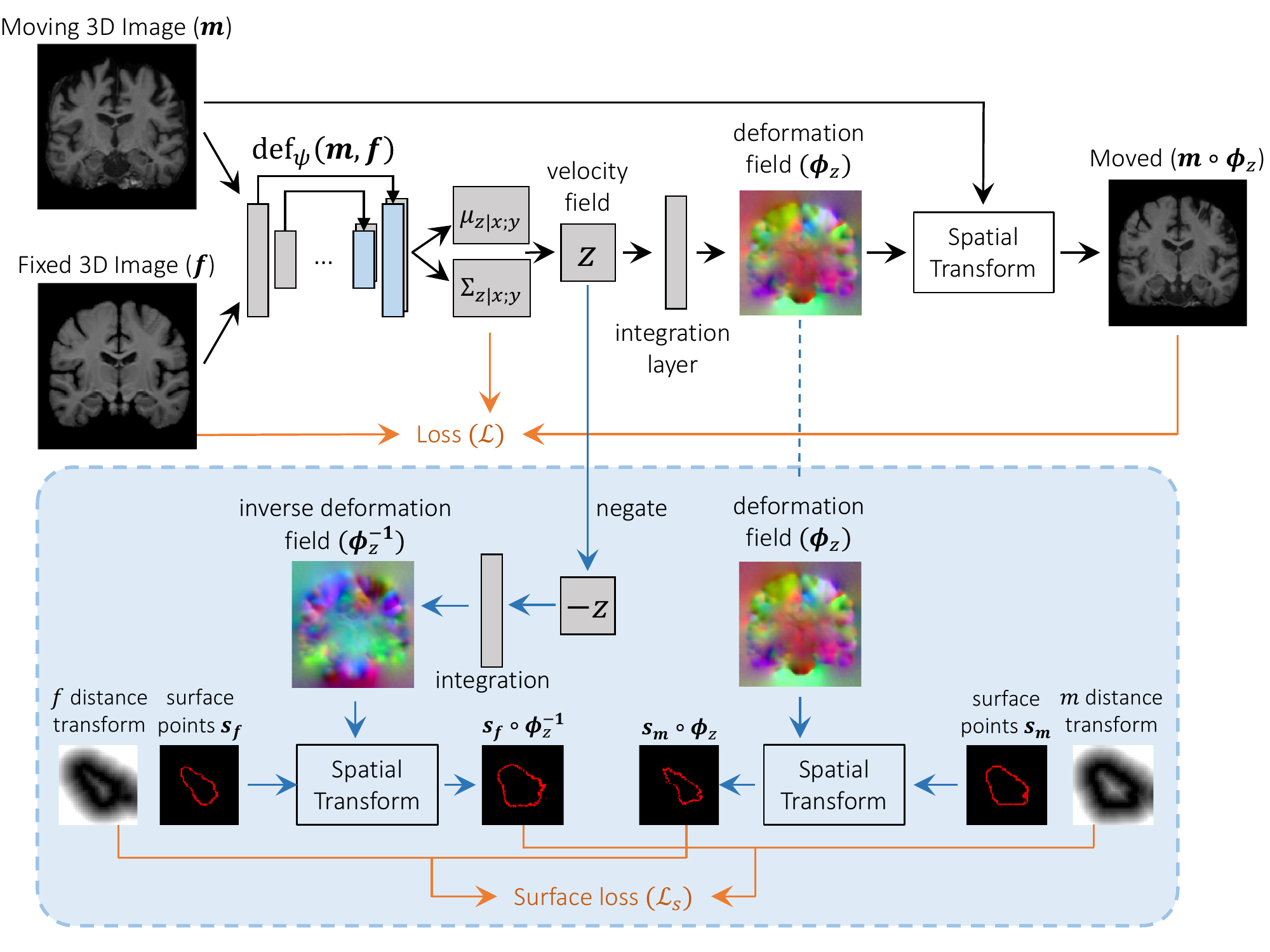}
	\end{center}
	\hfill
	\vspace{-0.75cm}
	\begin{minipage}[b]{1\linewidth}
		\caption{ Overview of end-to-end unsupervised architecture building on Figure~\ref{fig:network_overview_simple}. The first part of the network,~$\text{def}_{\psi}(\bmoving, \bfixed)$ takes the input images and outputs the approximate posterior probability parameters representing the velocity field mean,~$\bmu_{z|\moving;\fixed}$, and variance,~$\bSigma_{z|\moving;\fixed}$. A velocity field~$\bz$ is sampled and transformed to a diffeomorphic deformation field~$\bphi_z$ using novel differentiable \textit{squaring and scaling} layers. Finally, a spatial transform warps~$\bmoving$ to obtain~$\bmoving \circ \bphi_z$. The blue window illustrated the computation of \textit{optional} surface registration loss. The surface points and distance transform are computed for the both the moving and fixed surfaces. The surface points are warped by the resulting deformation, and a distance is computed using distance transforms.
		}
		\label{fig:sup:network_overview_simple}
	\end{minipage}
\end{figure*}

\clearpage
\twocolumn

\section*{Additional Figures}

\newpage

\begin{figure}[h!]
	\vspace{1.1cm}
	\begin{center}
		\includegraphics[width=\linewidth]{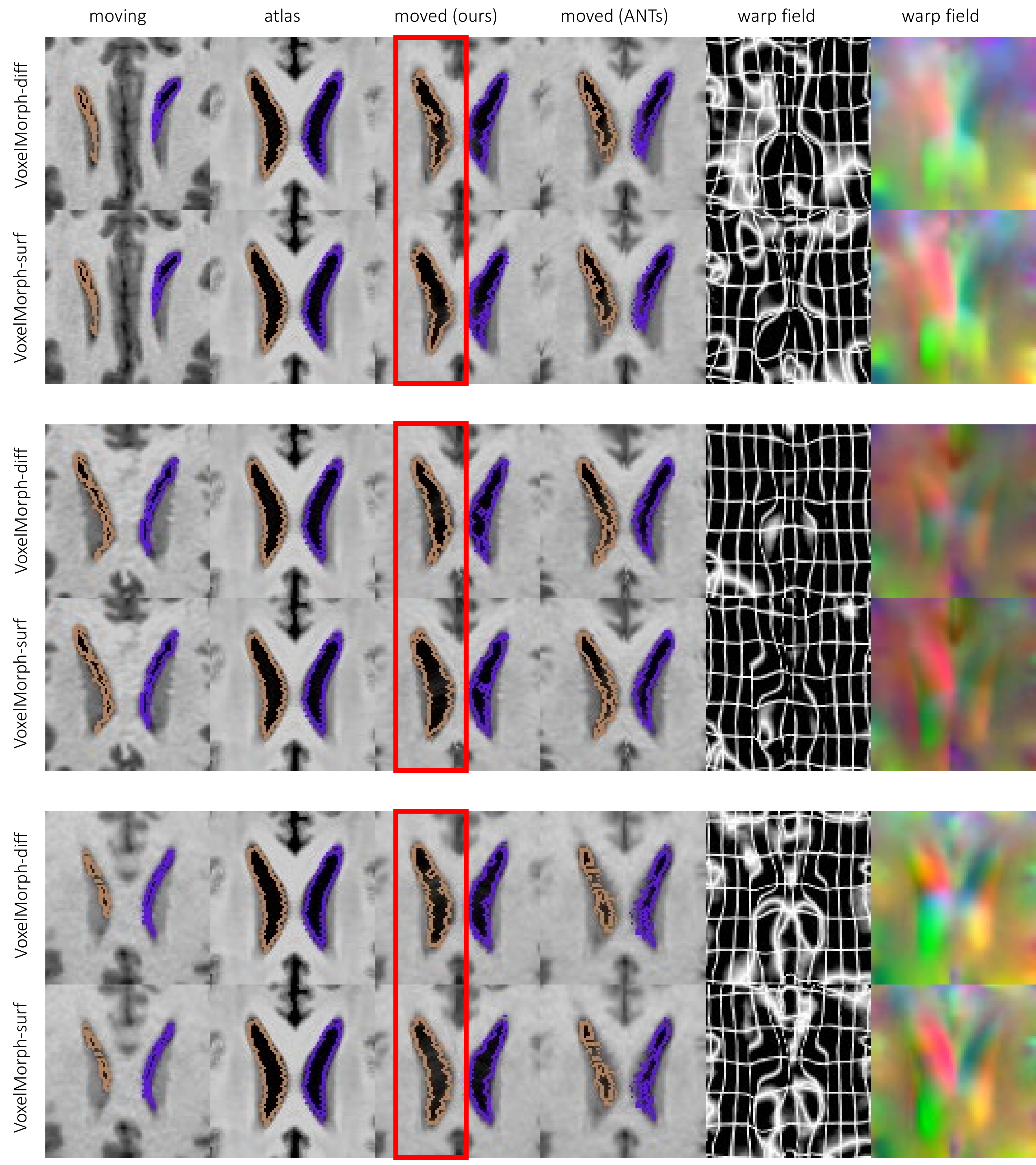}
	\end{center}
	\caption{Example surface-driven results. For three subjects, we show cropped MR slices of input moving image, atlas, and resulting warped image for our method and ANTs, with overlaid boundaries of ventricles for VoxelMorph-diff (top) and VoxelMorph-surf (bottom). For each set of two rows, we highlight in the red box an improvement in the segmentation of the ventricle from the top row to the bottom row.
	}
	\label{fig:sup:reg_examples_supervised}
\end{figure}

\end{document}